\documentclass[10pt,journal,cspaper,compsoc]{IEEEtran}
\ifCLASSOPTIONcompsoc
  \usepackage[nocompress]{cite}
\else
\fi
\ifCLASSINFOpdf
    \usepackage[pdftex]{graphicx}
    \DeclareGraphicsExtensions{.pdf,.jpeg,.png}
\else
    \usepackage[dvips]{graphicx}
    \DeclareGraphicsExtensions{.eps}
\fi
\usepackage[cmex10]{amsmath}
\usepackage{algorithmic}
\usepackage{url}

\usepackage{multirow}
\usepackage{algorithm}

\usepackage{color}
\usepackage{amsmath}
\usepackage{amssymb}
\usepackage{booktabs}
\usepackage{comment}

\begin{document}
%
\title{Explanatory Graphs for CNNs}

%
%

\author{Quanshi Zhang, Xin Wang, Ruiming Cao, Ying Nian Wu, Feng Shi, and Song-Chun Zhu, \textit{Fellow, IEEE}
\IEEEcompsocitemizethanks{\IEEEcompsocthanksitem Quanshi Zhang is with the Shanghai Jiao Tong University, Shanghai, China. Ruiming Cao, Feng Shi, Ying Nian Wu, and Song-Chun Zhu are with the University of California, Los Angeles, USA}
}

%
%

\markboth{IEEE TRANSACTIONS ON PATTERN ANALYSIS AND MACHINE INTELLIGENCE, in submission}%
{Shell \MakeLowercase{\textit{et al.}}: Bare Demo of IEEEtran.cls for Computer Society Journals}
%


\IEEEcompsoctitleabstractindextext{%

\begin{abstract}
This paper introduces a graphical model, namely an explanatory graph, which reveals the knowledge hierarchy hidden inside conv-layers of a pre-trained CNN. Each filter\footnote[1]{The output of a conv-layer is called the feature map of a conv-layer. Each channel of this feature map is produced by a filter, so we call a channel the feature map of a filter.} in a conv-layer of a CNN for object classification usually represents a mixture of object parts. We develop a simple yet effective method to disentangle object-part pattern components from each filter. We construct an explanatory graph to organize the mined part patterns, where a node represents a part pattern, and each edge encodes co-activation relationships and spatial relationships between patterns. More crucially, given a pre-trained CNN, the explanatory graph is learned without a need of annotating object parts. Experiments show that each graph node consistently represented the same object part through different images, which boosted the transferability of CNN features. We transferred part patterns in the explanatory graph to the task of part localization, and our method significantly outperformed other approaches.
\end{abstract}


}

\maketitle
\IEEEdisplaynotcompsoctitleabstractindextext
\IEEEpeerreviewmaketitle

\section{Introduction}

Convolutional neural networks (CNNs)~\cite{CNN,CNNImageNet,ResNet} have exhibited superior performance in various visual tasks, for example, object classification and detection. In comparison, explaining features in middle conv-layers of a CNN has presented continuous challenges for decades. When a CNN is trained for object classification, its conv-layers have encoded rich implicit patterns of object parts and patterns of textures. Therefore, this research aims to provide a global analysis of how visual knowledge is organized in a pre-trained CNN:
\begin{itemize}
\item[1] How many patterns can activate a certain convolutional filter of the CNN? For example, the filter may be triggered by both a specific object-part pattern or a certain textural pattern.
\item[2] Which patterns are co-activated to describe an object part?
\item[3] What is the spatial relationship between two co-activated patterns?
\end{itemize}

Given a CNN pre-trained for object classification, in this paper, we propose a method (i) to mine object-part patterns from intermediate conv-layers and (ii) to organize these patterns in an explanatory graph.

\begin{figure*}[t]
\centering
\includegraphics[width=\linewidth]{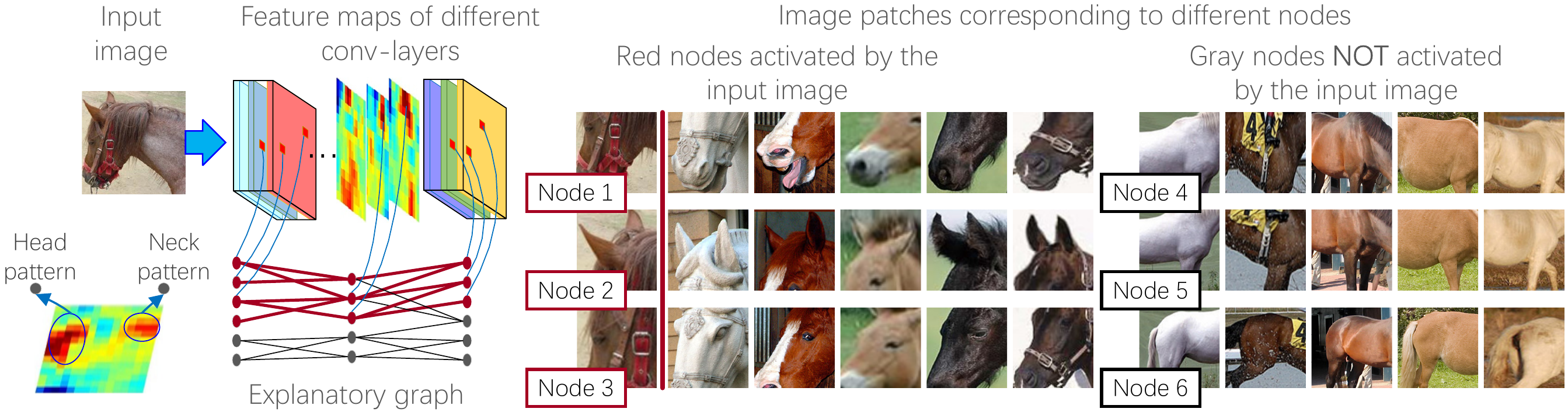}
\caption{An explanatory graph represents knowledge hierarchy hidden in conv-layers of a CNN. Each filter in a pre-trained CNN may be activated by different object parts. Our method disentangles part patterns from each filter in an unsupervised manner, thereby clarifying the knowledge representation.}
\label{fig:top}
\end{figure*}

As shown in Fig.~\ref{fig:top}, the explanatory graph encodes the knowledge hierarchy hidden inside the CNN, as follows.
\begin{itemize}
\item The explanatory graph has multiple layers, which correspond to different conv-layers of the CNN.
\item Each graph layer has many nodes. We use graph nodes in a layer to represent all candidate part patterns that can activate the feature map of the corresponding conv-layer.
\item Because a filter in the conv-layer may be potentially triggered by multiple parts of the object, we disentangle different part patterns from the same filter, which are represented as different graph nodes.
\item A graph edge connects two nodes in adjacent layers to encode co-activation logics and spatial relationships between them.
\item We can regard the explanatory graph as a dictionary, which summarizes the part knowledge hidden inside hundreds of thousands of chaotic neural activations of a conv-layer into thousands of graph nodes.
\item During the inference process, given feature maps of an input image, our method selects a small number of nodes from the explanatory graph and assigns these nodes with certain neural activations in the feature map. We consider these nodes are activated to explain which part patterns are hidden behind the neural activations. Each graph node consistently corresponds to the same part over different input object images.
\end{itemize}
Note that the location of each pattern (node) is not fixed to a specific neural activation unit during the inference process. Instead, given different input images, a part pattern may appear on various locations of a filter's feature maps\footnotemark[1]. For example, the ear pattern and the face pattern of a horse in Fig.~\ref{fig:pair} can appear on different locations of different images, but they are co-activated and keep certain spatial relationships.

\noindent\textbf{\textbullet\quad Disentangling object parts from a single filter} is the core technique of building an explanatory graph. As shown in Fig.~\ref{fig:top}, a filter in a conv-layer may be activated by different object parts (\emph{e.g.} the filter's feature map\footnotemark[1] may be activated by both the head and the neck of a horse).

In this study, we hope to develop a simple yet effective method to automatically disentangle different part patterns from a single filter without using any annotations of object parts, which presents considerable challenges for state-of-the-art algorithms. In this way, the explanatory graph explains neural activations with clear meanings and ignores noisy activations and activations of textural patterns. Given a testing image to the CNN, the explanatory graph can infer (i) which nodes (parts) are responsible for neural activations of a filter and (ii) locations of the corresponding parts on the feature map.

\noindent\textbf{\textbullet\quad Graph nodes with high transferability:} The explanatory graph contains off-the-shelf patterns for object parts. The explanatory graph summarizes chaotic feature maps of conv-layers into object parts, which can be considered as a more concise and meaningful representation of the CNN knowledge, just like a dictionary. The explanatory graph enables us to accurately transfer object-part patterns from conv-layers to other tasks. Because all filters in the CNN are learned using numerous training images, we can consider each graph node as a detector that has been sophisticatedly learned to detect a part among thousands of images.

To prove the above assertions, we learn explanatory graphs for different CNNs (including the VGG-16, residual networks, and the encoder of a VAE-GAN) and analyze the graphs from various perspectives as follows.\\
\textit{Visualization \& reconstruction:} We visualize part patterns encoded by graph nodes using the following two approaches. First, for each graph node, we select object parts that most strongly activate the node for visualization. Second, we learn another decoder network to invert activation states of graph nodes to reconstruct image regions of the nodes.\\
\textit{Examining part interpretability of graph nodes:} We quantitatively evaluate the part-level interpretability of graph nodes. Given an explanatory graph, we measure whether a node consistently represents the same part on different objects.\\
\textit{Examining location instability of graph nodes:} Besides the part interpretability, we also use a new metric, namely location instability, to measure the semantic clarity of each graph node. It is assumed that if a graph node consistently represents the same object part, then the distance between the inferred part and some ground-truth landmarks of the object should not change a lot through different images. Thus, the evaluation metric uses the deviation of such relative distances over images to measure the instability of a part pattern.\\
\textit{Testing transferability:} The transferability of graph nodes are tested in the scenario of few-shot part localization. We associate certain graph nodes with explicit part names based on feature maps of very few images, in order to localize the target part. The superior localization performance proves the good transferability of graph nodes.

\textbf{Contributions} of this paper are summarized as follows.
\begin{itemize}
\item In this paper, we, for the first time, propose a simple yet effective method to extract and summarize part knowledge hidden inside chaotic feature maps of intermediate conv-layers of a CNN and organize the layerwise knowledge hierarchy using an explanatory graph. Experiments show that each graph node consistently represents the same object part through different input images.
\item As a generic method, we can learn explanatory graphs for different CNNs, \emph{e.g.} VGGs, residual networks, and the encoder of a VAE-GAN.
\item Graph nodes (patterns) have good transferability, especially in the task of few-shot part localization. Although our graph nodes were learned without part annotations, our transfer-learning-based part localization still outperformed approaches that learned part representations using part annotations.
\end{itemize}
A preliminary version of this paper appeared in \cite{explanatoryGraph}.

\section{Related work}

\subsection{Semantics in CNNs}

The interpretability and the discrimination power are two crucial aspects of a CNN~\cite{Interpretability}. In recent years, different methods are developed to explore the semantics hidden inside a CNN.

\textbf{Visualization \& interpretability of CNN filters:}{\verb| |} Visualization of filters in a CNN is the most direct way of exploring the pattern hidden inside a neural unit. Lots of visualization methods have been used in the literature. Dosovitskiy~\emph{et al.}~\cite{FeaVisual} proposed up-convolutional nets to invert feature maps of conv-layers to images. However, up-convolutional nets cannot mathematically ensure the visualization result reflects actual neural representations. Comparatively, gradient-based visualization~\cite{CNNVisualization_1,CNNVisualization_2,CNNVisualization_3} showed the appearance that maximized the score of a given unit, which is more close to the spirit of understanding CNN knowledge. Furthermore, Bau \emph{et al.}~\cite{Interpretability} defined and analyzed the interpretability of each filter. In recent years, \cite{olah2017feature} provided a reliable tool to visualize filters in different conv-layers of a CNN.

Although these studies achieved clear visualization results, theoretically, gradient-based visualization methods visualize one of the local minimums contained in a high-layer filter. \emph{I.e.} when a filter represents multiple patterns, these methods selectively illustrated one of the patterns; otherwise, the visualization result will be chaotic. Similarly, \cite{Interpretability} selectively analyzed the semantics among the highest 0.5\% activations of each filter. In contrast, our method provides a solution to explaining both strong and relatively weak activations from each filter, instead of exclusively extracting significant neural activations.

\textbf{Active network diagnosis:}{\verb| |} Going beyond ``passive'' visualization, some methods ``actively'' diagnose a pre-trained CNN to obtain insight understanding of CNN representations. Many statistical methods~\cite{CNNAnalysis_1,CNNAnalysis_2,CNNVisualization_5} have been proposed to analyze the characteristics of CNN features. \cite{CNNAnalysis_1} explored semantic meanings of convolutional filters. \cite{CNNAnalysis_2} evaluated the transferability of filters in intermediate conv-layers. \cite{CNNAnalysis_3,CNNVisualization_5} computed feature distributions of different categories in the CNN feature space. Methods of \cite{visualCNN_grad,visualCNN_grad_2} propagated gradients of feature maps \emph{w.r.t.} the CNN loss back to the image, in order to estimate the image regions that directly contribute the network output. The LIME~\cite{trust} and the SHAP~\cite{shap} proposed general methods extract input units of a neural network that are used for a specific prediction.

Zhang~\emph{et al.}~\cite{CNNBias} has demonstrated that in spite of the good classification performance, a CNN may encode biased knowledge representations due to dataset bias. Instead, the CNN usually uses unreliable contexts for classification. For example, a CNN may extract features from hairs as a context to identify the \textit{smiling} attribute.

Therefore, in order to ensure the correctness of feature representations, network-attack methods~\cite{pixelAttack,CNNInfluence,CNNAnalysis_1} diagnosed network representations by computing adversarial samples for a CNN. In particular, influence functions~\cite{CNNInfluence} were proposed to compute adversarial samples, provide plausible ways to create training samples to attack the learning of CNNs, fix the training set, and further debug representations of a CNN. \cite{banditUnknown} discovered knowledge blind spots (unknown patterns) of a pre-trained CNN in a weakly-supervised manner. Some studies~\cite{wu2007compositional,yang2009evaluating,wu2011numerical} mined the local, bottom-up, and top-down information components in a model to construct a hierarchical object representation. From this perspective, our method disentangles object-part patterns from a pre-trained CNN and builds a knowledge hierarchy to diagnose the knowledge inside the CNN.

\textbf{Pattern retrieval:}{\verb| |} Some studies retrieve units with specific meanings from CNNs for different applications. Like middle-level feature extraction~\cite{MiddleLevel}, pattern retrieval mainly learns mid-level representations of CNN knowledge. Zhou~\emph{et al.}~\cite{CNNSemanticDeep,CNNSemanticDeep2} selected units from feature maps to describe ``scenes''. In particular, \cite{CNNSemanticDeep} proposed a method to accurately compute the image-resolution receptive field of neural activations in a feature map. Theoretically, the actual receptive field of a neural activation is smaller than that computed using the filter size. The accurate estimation of the receptive field is crucial to understand a filter's representations. Simon~\emph{et al.} discovered objects from feature maps of unlabeled images~\cite{ObjectDiscoveryCNN_2}, and selected a filter to describe each part in a supervised fashion~\cite{CNNSemanticPart}. However, most methods simply assumed that each filter mainly encoded a single visual concept, and ignored the case that a filter in high conv-layers encoded a mixture of patterns. \cite{CNNAoG,DeepQA,holdingHands} extracted certain neurons from a filter's feature map to describe an object part in a weakly-supervised manner (\emph{e.g.} learning from active question answering and human interactions).

\begin{figure*}[t]
\centering
\includegraphics[width=\linewidth]{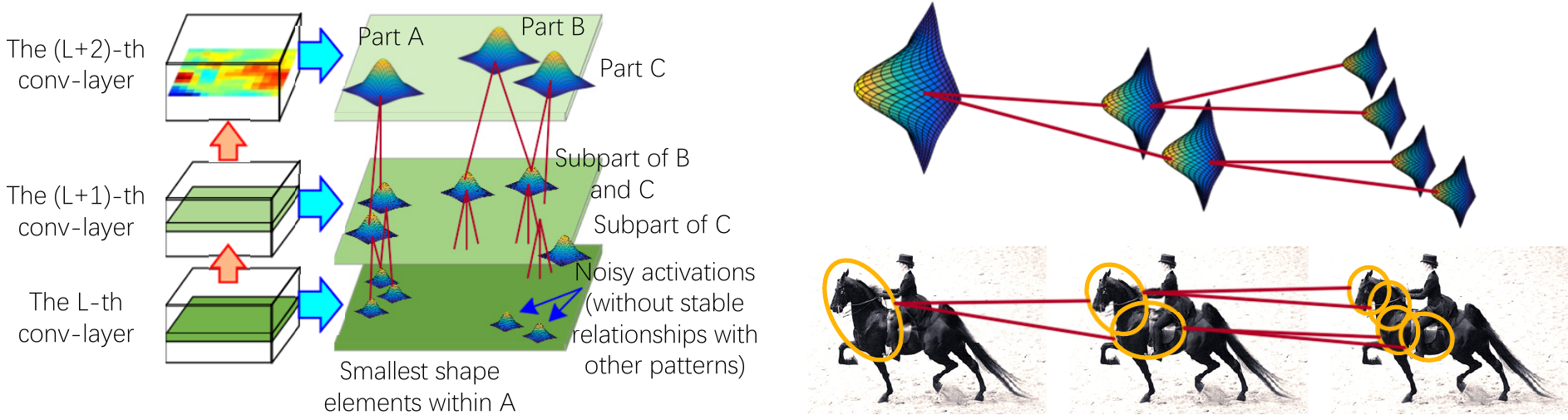}
\caption{Schematic illustration of the explanatory graph. The explanatory graph encodes spatial and co-activation relationships between part patterns in the explanatory graph. High-layer patterns filter out noises and disentangle low-layer patterns. From another perspective, we can regard low-layer patterns as components of high-layer patterns.}
\label{fig:peak}
\end{figure*}

In this study, the explanatory graph disentangles patterns different parts in the CNN without a need of part annotations. Compared to raw feature maps, patterns in graph nodes are more interpretable.

\textbf{CNN semanticization:}{\verb| |} Compared to the diagnosis of CNN representations and the pattern retrieval, semanticization of CNN representations is closer to the spirit of building interpretable representations.

Hu~\emph{et al.}~\cite{LogicRuleNetwork} designed logic rules for network outputs, and used these rules to regularize neural networks and learn meaningful representations. However, this study has not obtained semantic representations in intermediate layers. \cite{additiveExplainer} distilled knowledge of a neural network into an additive model to explain the knowledge inside the network. \cite{explanatoryTree_arXiv} used a tree structure to summarize the inaccurate rationale of each CNN prediction into generic decision-making models for a number of samples. Capsule nets~\cite{capsule} and interpretable CNNs~\cite{interpretableCNN} used certain network structures and loss functions, respectively, to make the network automatically encode interpretable features in intermediate layers.

In comparison, we aim to explore the entire semantic hierarchy hidden inside conv-layers of a CNN. With clear semantic structures, the explanatory graph makes it easier to transfer CNN patterns to other part-based tasks.

\subsection{Weakly-supervised knowledge transferring}

Knowledge transferring ideas have been widely used in deep learning. Typical research includes end-to-end fine-tuning and transferring CNN knowledge between different categories~\cite{CNNAnalysis_2} or different datasets~\cite{UnsuperTransferCNN}. In contrast, a transparent representation of part knowledge will create a new possibility of transferring part knowledge to other applications. Therefore, we build an explanatory graph to represent part patterns hidden inside a CNN, which enables transfer part patterns to other tasks. Experiments have demonstrated the efficiency of our method in few-shot part localization.

\section{Algorithm}

A single filter is usually activated by different parts of the object (see Fig.~\ref{fig:peak}). Let us assume that given an input image, a filter is activated by $N$ parts, \emph{i.e.} there are $N$ activation peaks on the filter's feature map. Some peaks represent common parts of the object, which are termed \textit{part patterns}. Other activation peaks may correspond to background noises or textural patterns.

Our goal is to disentangle activation peaks corresponding to part patterns from chaotic feature maps of a filter. It is assumed that if an activation peak of a filter represents an object part, then the CNN usually also contains other filters to represent neighboring parts of the target part. \emph{I.e.} some activation peaks (patterns) of these filters must keep certain spatial relationships with the target part. Thus, the explanatory graph connects each pattern (node) in a low layer to some patterns in the neighboring upper layer.

We mine part patterns layer by layer. Given patterns mined from the upper layer, we extract activation peaks that keep stable spatial relationships with specific upper-layer patterns through different images, as part patterns in the current layer.

Patterns in high layers usually represent large-scale object parts, while patterns in low layers mainly describe small and relatively simple shapes, which can be regarded as components of high-layer patterns. Patterns in high layers are usually discriminative, and the explanatory graph uses high-layer patterns to filter out noisy activations. Patterns in low layers are disentangled based on their spatial relationship with high-layer patterns.

\subsection{Learning}

We are given a CNN, which is pre-trained using its own set of training samples ${\bf I}$. Let $G$ denote the target explanatory graph. $G$ contains several layers corresponding to conv-layers in the CNN. Our method disentangles the $d$-th filter of the $L$-th conv-layer into $N_{L,d}$ part patterns. These part patterns are modeled as a set of $N_{L,d}$ nodes in the $L$-th layer of $G$, denoted by $\Omega_{L,d}$. $\Omega_{L}=\cup_{d}\Omega_{L,d}$ is given as the entire node set for the $L$-th layer. ${\boldsymbol\theta}_{L}$ represents parameters of nodes in the $L$-th layer, which mainly encode spatial relationships between these nodes and nodes in the $(L+1)$-th layer.

Given an input image {$I\in{\bf I}$}, the $L$-th conv-layer of the CNN generates a feature map\footnotemark[1], denoted by ${\bf X}_{L}^{I}$. Then, for each node $V\in\Omega_{L,d}$, the explanatory graph infers whether or not $V$'s part pattern appears on the $d$-th channel\footnotemark[1] of ${\bf X}_{L}^{I}$, as well as the part location (if the pattern appears). We use ${\bf R}_{L}^{I}$ to represent position inference results for all nodes in the $L$-th layer.

\begin{figure*}[t]
\centering
\includegraphics[width=0.8\linewidth]{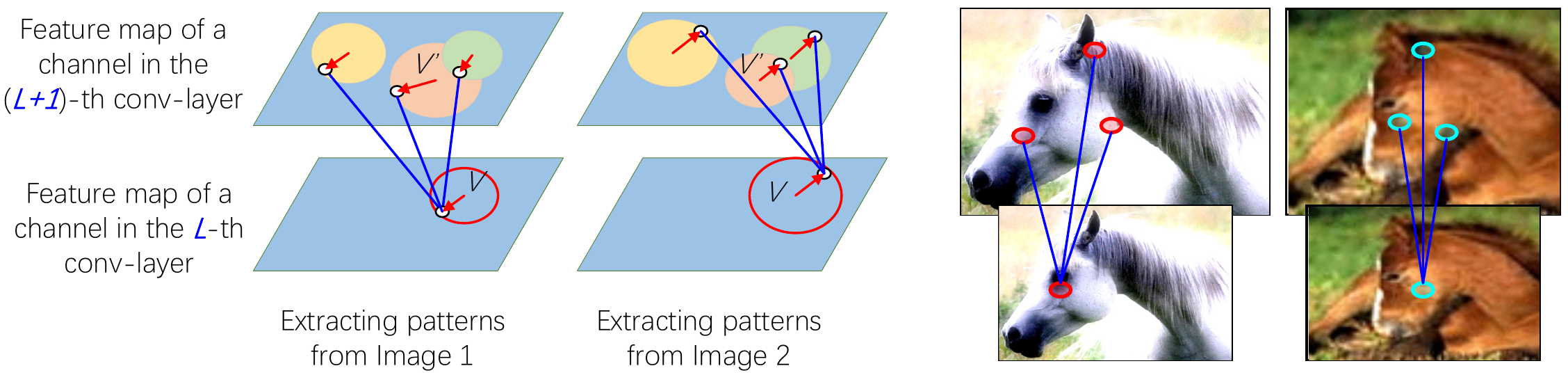}
\caption{Schematic illustration of related patterns $V$ and $V'$. The related patterns keep similar spatial relationships among different images. Circle centers represent the prior pattern positions, \emph{e.g.} $\mu_{V}$ and $\mu_{V'}$. Red arrows denote relative displacements between the inferred positions and prior positions, \emph{e.g.} ${\bf p}_{V}-\mu_{V}$.}
\label{fig:pair}
\end{figure*}

\textbf{Top-down iterative learning of explanatory graphs:}{\verb| |} Given all training images ${\bf I}$, we expect that (i) all patterns nodes in the explanatory graph can be well fit to feature maps of all images, and (ii) nodes in the lower layer always keep consistent with nodes in the upper layer given each input images. Therefore, the learning of an explanatory graph is conducted in a top-down manner as follows.

We first disentangle patterns from the top conv-layer of the CNN and construct the top graph layer. Then, we use inference results of the patterns/nodes on the top layer to help disentangle patterns from the neighboring lower conv-layer. In this way, we can ensure stable layerwise spatial relationships between patterns.

When we learn the $L$-th layer, for each node $V$, we need to learn the following two terms: (i) the parameter $\mu_{V}\in{\boldsymbol\theta}_{L}$ and (ii) a set of patterns in the upper layer that are connected to $V$, ${E}_{V}\in{\boldsymbol\theta}_{L}$. $\mu_{V}\in{\boldsymbol\theta}_{L}$ denotes the prior location of $V$. Thus, for each node $V'\in{E}_{V}$, $\mu_{V}-\mu_{V'}$ corresponds the prior displacement between $V$ and the upper node $V'$. The explanatory graph only uses the displacement $\mu_{V}-\mu_{V'}$ to model the spatial relationships between nodes.

Just like an EM algorithm, we use the current explanatory graph to fit feature maps of training images. Then, we use matching results as feedback to modify the prior location $\mu_{V}$ and edge connections ${E}_{V}$ of each node $V$ in the $L$-th layer, in order to make the explanatory graph better fit the feature maps. We repeat this process iteratively to obtain the optimal prior location and edge connections for $V$.

In other words, our method automatically extracts pairs of related patterns and learns the optimal spatial relationships between them during the iterative learning process, which best fit feature maps of training images.

Therefore, the objective function of learning the $L$-th layer is given as
\begin{equation}
{\arg\!\max}_{{\boldsymbol\theta}_{L}}{\prod}_{I\in{\bf I}}P({\bf X}_{L}^{I}|{\bf R}_{L+1}^{I},{\boldsymbol\theta}_{L})
\label{eqn:prob}
\end{equation}
Let us focus on the feature map {${\bf X}_{L}^{I}$} of image $I$. Without ambiguity, we ignore the superscript $I$ to simplify notations in following paragraphs. We can regard {${\bf X}_{L}$} as a distribution of ``neural activation entities.'' The neural response of each unit {$x\in{\bf X}_{L}$} can be considered as the number of ``activation entities.'' In other words, each neural unit $x$ localizes at the position of ${\bf p}_{x}$\footnote[2]{To make unit positions in different conv-layers comparable with each other (\emph{e.g.} {$\mu_{V'\rightarrow V}$} in Eq.~\ref{eqn:gauss}), we project the position of unit $x$ to the image plane. We define the coordinate {${\bf p}_{x}$} on the image plane, instead of on the feature-map plane.} in the $d_{x}$-th channel of ${\bf X}_{L}$. We use $F(x)=\beta\cdot\max\{f_{x},0\}$ to denote the number of activation entities at the location ${\bf p}_{x}$, where $f_{x}$ is the normalized response value of $x$; $\beta$ is a constant.

Just like a Gaussian mixture model, all patterns in $\Omega_{L,d}$ comprise a mixture model, which explains the distribution of activation entities on the $d$-th channel of {${\bf X}_{L}$}. Each node {$V\in\Omega_{L,d}$} is treated as a hidden variable or an alternative component in the mixture model to describe activation entities.
\begin{small}
\begin{eqnarray}
\!\!\!\!&P({\bf X}_{L}|{\bf R}_{L+1},{\boldsymbol\theta}_{L})\!=\!{\prod}_{x\in{\bf X}_{L}}P({\bf p}_{x}|{\bf R}_{L+1},{\boldsymbol\theta}_{L})^{F(x)}\\
\!\!\!\!&={\prod}_{x\in{\bf X}_{L}}\Big\{\sum\limits_{V\in\Omega_{L,d}\cup\{V_{\textrm{none}}\}}\!\!\!\!\!\!\!\!\!P(V)P({\bf p}_{x}|V,{\bf R}_{L+1},{\boldsymbol\theta}_{L})\!\Big\}_{d=d_{x}}^{F(x)}\!\nonumber
\end{eqnarray}
\end{small}
where $P(V)=\frac{1}{N_{L,d}+1}$ is a constant prior probability. $P({\bf p}_{x}|V,{\bf R}_{L+1},{\boldsymbol\theta}_{L})$ measures the compatibility of using node $V$ to describe an activation entity at ${\bf p}_{x}$. In particular, we add a dummy component $V_{\textrm{none}}$ to the mixture model for noisy activations, which cannot be explained by any part patterns. The compatibility between $V$ and ${\bf p}_{x}$ is based on spatial relationship between $V$ and its connected nodes in $G$, which is approximated as
\begin{small}
\begin{eqnarray}
\!\!\!\!P({\bf p}_{x}|V,{\bf R}_{L+1},{\boldsymbol\theta}_{L})\!=\!\left\{\!\begin{array}{ll}\gamma\!\!\!\!\!\prod\limits_{V'\in{E}_{V}}\!\!\!P({\bf p}_{x}|{\bf p}_{V'},{\boldsymbol\theta}_{L})^{\lambda}\!\!,\!\!\!&\!\!\!V\!\in\!\Omega_{L\!,d_{x}}\!\!\!\!\!\!\!\!\!\!\!\!\!\!\\
\gamma\tau,&\!\!\!\!\!\!\!\!\!\!\!\!V\!=\!V_{\textrm{none}}\!\!\!\!\!\!\!\!\!\!\!\!\\
\end{array}\right.\label{eqn:prob-full}\\
\!\!\!P({\bf p}_{x}|{\bf p}_{V'},{\boldsymbol\theta}_{L})\!=\!{\bf\mathcal N}({\bf p}_{x}|\mu_{V'\!\rightarrow\!V},\sigma_{V'}^2)\!\!\label{eqn:gauss}
\end{eqnarray}
\end{small}
In above equations, $V$ has $M$ related nodes in the upper layer. The set of node connections ${E}_{V}\in{\boldsymbol\theta}_{L}$ would be determined during the learning process. The overall compatibility {$P({\bf p}_{x}|V,{\bf R}_{L+1},{\boldsymbol\theta}_{L})$} is divided into the spatial compatibility between node $V$ and each related node $V'$, $P({\bf p}_{x}|{\bf p}_{V'},{\boldsymbol\theta}_{L})$. $\forall V'\in{E}_{V}$, ${\bf p}_{V'}\!\in\!{\bf R}_{L+1}$ denotes the position inference result of $V'$, which have been given. $\lambda=\frac{1}{M}$ is a constant for normalization. $\gamma$ is a constant to roughly ensure $\int P({\bf p}_{x}|V,{\bf R}_{L+1},{\boldsymbol\theta}_{L}){\bf d}{{\bf p}_{x}}=1$, which can be eliminated during the learning process.

As shown in Fig.~\ref{fig:pair}, an intuitive idea is that the relative displacement between $V$ and $V'$ should not change a lot among different images. Then, {${\bf p}_{x}-{\bf p}_{V'}$} will approximate to the prior displacement $\mu_{V}-\mu_{V'}$, if node $V$ can well fit the activation at ${\bf p}_{x}$. Given ${E}_{V}$, we assume the spatial relationship between $V$ and $V'$ follows a Gaussian distribution in Eqn.~\ref{eqn:gauss}, where we define $\mu_{V'\rightarrow V}=\mu_{V}-\mu_{V'}+{\bf p}_{V'}$ as the prior localization of $V$ given $V'$. The variation $\sigma_{V'}^2$ can be estimated from data\footnote[3]{We can prove that for each $V\in\Omega_{L,d}$, $P({\bf p}_{x}|V,{\bf R}_{L+1},{\boldsymbol\theta}_{L})$ $\propto{\bf\mathcal N}({\bf p}_{x}|\mu_{V}$ $+\Delta_{I,V},\tilde{\sigma}_{V}^2)$, where $\Delta_{I,V}=\sum_{V'\in{E}_{V}}$ $\frac{{\bf p}_{V'}-\mu_{V'}}{\sigma_{V'}^2}$ $/\sum_{V'\in{E}_{V}}\frac{1}{\sigma_{V'}^2}$; $\tilde{\sigma}_{V}^2$ $=1/{\bf E}_{V'\in{E}_{V}}\frac{1}{\sigma_{V'}^2}$. Therefore, we can either directly use $\tilde{\sigma}_{V}^2$ as $\sigma_{V}^2$, or compute the variation of ${\bf p}_{x}-\mu_{V}-\Delta_{I,V}$ \emph{w.r.t.} different images to obtain $\sigma_{V}^2$.}.

\begin{algorithm}[t]
\begin{algorithmic}
\STATE {\bf Inputs:} feature map ${\bf X}_{L}$ of the $L$-th conv-layer, inference results ${\bf R}_{L+1}$ in the upper conv-layer.
\STATE {\bf Outputs:} $\mu_{V},{E}_{V}$ for $\forall V\in\Omega_{L}$.
\STATE {\bf Initialization:} $\forall V$, ${E}_{V}\!=\!\{V_{\textrm{dummy}}\}$, a random value for $\mu_{V}^{(0)}$
\FOR{$iter=1$ to $T$}
\STATE $\forall V\in\Omega_{L}$, compute $P({\bf p}_{x},V|{\bf R}_{L+1},{\boldsymbol\theta}_{L})$.
\FOR{$V\in\Omega_{L}$}
\STATE Update $\mu_{V}$ via an EM algorithm,\\
{\small$\mu_{V}^{(iter)}\!=\!\mu_{V}^{(iter-1)}\!\!+\!\eta\!\!\!\!\!\!\!\sum\limits_{I\in{\bf I},x\in{\bf X}_{L}}\!\!\!\!\!\!\!{\bf\large E}_{P(V|{\bf p}_{x},{\bf R}_{L+1},{\boldsymbol\theta}_{L})}\big[$ $F(x)\cdot\frac{\partial{\log}P({\bf p}_{x},V|{\bf R}_{L+1},{\boldsymbol\theta}_{L})}{\partial\mu_{V}}\big]$}.\\
\STATE  Select $M$ patterns from $V'\in\Omega_{L+1}$ to construct ${E}_{V}$ based on a greedy strategy, which maximize {\small${\prod}_{I\in{\bf I}}P({\bf X}_{L}|{\bf R}_{L+1},{\boldsymbol\theta}_{L})$}.
\ENDFOR
\ENDFOR
\end{algorithmic}
\caption{Learning sub-graph in the $L$-th layer}
\label{alg:main}
\end{algorithm}

We learn the explanatory graph in a top-down manner, and the learning process is summarized in Algorithm~\ref{alg:main}. We first learn nodes in the top-layer of $G$, and then learn for the neighboring lower layer. For the sub-graph in the $L$-th layer, we iteratively estimate $\mu_{V}$ and ${E}_{V}$ for nodes in the sub-graph.

Note that for each pattern $V$ in the top conv-layer, we simply define ${E}_{V}\!=\!\{V_{\textrm{dummy}}\}$, $\mu_{V_{\textrm{dummy}}}={\bf p}_{V_{\textrm{dummy}}}={\bf 0}$, where $V_{\textrm{dummy}}$ is a node in the dummy layer above the top conv-layer. Based on Eqns.~(\ref{eqn:prob-full}) and (\ref{eqn:gauss}), we obtain $P({\bf p}_{x}|V,{\bf R}_{L+1},{\boldsymbol\theta}_{L})={\bf\mathcal N}({\bf p}_{x}|\mu_{V},\sigma_{V}^2)$.

\textbf{Inference of pattern locations:} Given feature maps of an input image, we can assign nodes in the explanatory graph with different activations peaks on feature maps, in order to infer semantic meanings (parts) represented by these neural activations. The explanatory graph simply assigns node $V\in\Omega_{L,d}$ with the unit $\hat{x}={\arg\!\max}_{x\in{\bf X}_{L}:d_{x}=d}S_{V\rightarrow x}^{I}$ on the feature map as the inference of $V$, where $S_{V\rightarrow x}^{I}\!=\!F(x)P({\bf p}_{x}|V,{\bf R}_{L+1},{\boldsymbol\theta}_{L})$ denotes the score of assigning $V$ to $x$. Accordingly, ${\bf p}_{V'}={\bf p}_{\hat{x}}$ represents the inferred location of $V$. In particular, in Eqn.~(\ref{eqn:prob}), we define ${\bf R}_{L+1}=\{{\bf p}_{V'}\}_{V'\in\Omega_{L+1}}$.

\section{Experiments}

To demonstrate the broad applicability of our method, we learned explanatory graphs to interpret four types of CNNs, \emph{i.e.} the VGG-16~\cite{VGG}, the 50-layer and 152-layer Residual Networks~\cite{ResNet}, and the encoder of the VAE-GAN~\cite{VAEGAN}. These CNNs learned using a total of 37 animal categories in three datasets, which included the ILSVRC 2013 DET Animal-Part dataset~\cite{CNNAoG}, the CUB200-2011 dataset~\cite{CUB200}, and the VOC Part dataset~\cite{SemanticPart}. As discussed in \cite{SemanticPart,CNNAoG}, animals usually contain non-rigid parts, which presents a key challenge for part localization. Thus, we selected animal categories in the three datasets for testing.

We designed three experiments to evaluate the explanatory graph from different perspectives. In the first experiment, we visualized node patterns in the explanatory graph. The second experiment was designed to evaluate the interpretability of part patterns, \emph{i.e.} checking whether or not a node pattern consistently represents the same object part among different images. We compared our patterns with three types of middle-level features and neural patterns. In the third experiment, we used our graph nodes for the task of few-shot part localization, in order to test the transferability of node patterns in the graph. We associated part patterns with explicit part names for part localization. We compared our part-localization performance with fourteen baselines.

\subsection{Implementation details}

\begin{figure}[t]
\centering
\includegraphics[width=0.6\linewidth]{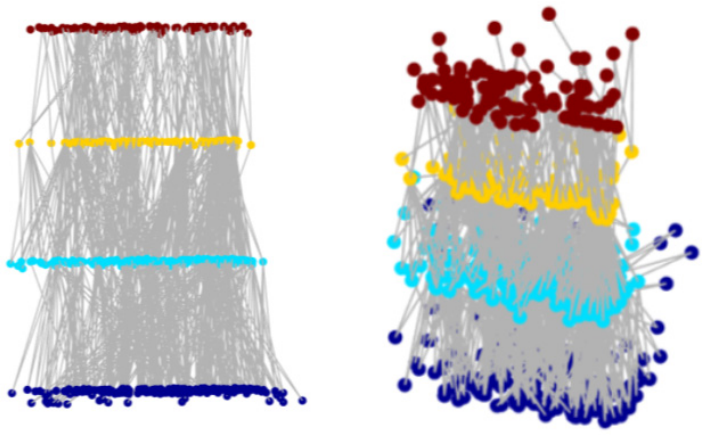}
\caption{A four-layer explanatory graph. For clarity, we put all nodes of different filters in the same conv-layer on the same plan and only show 1\% of the nodes with 10\% of their edges from two perspectives.}
\label{fig:global}
\end{figure}

\begin{figure*}[t]
\centering
\includegraphics[width=\linewidth]{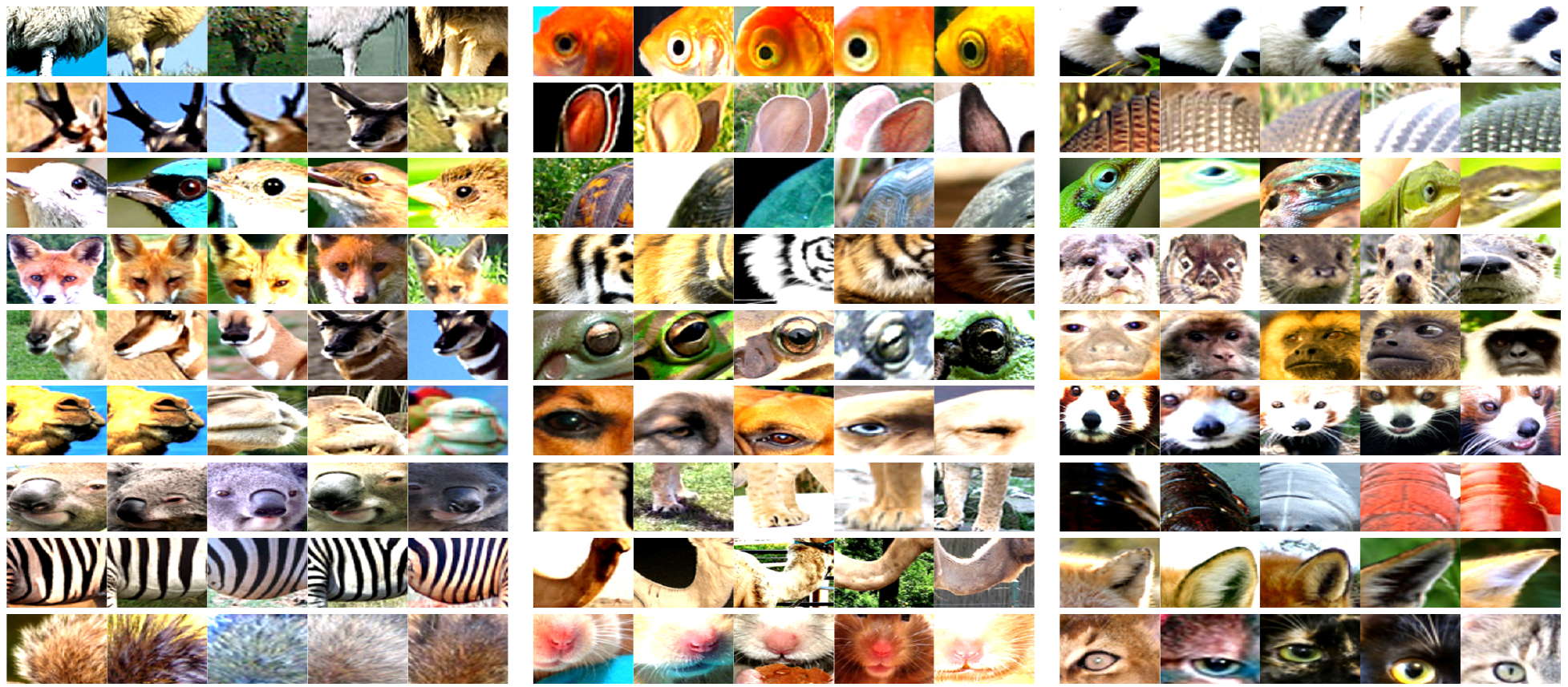}
\caption{Image patches corresponding to different nodes in explanatory graphs.}
\label{fig:patch}
\end{figure*}

\begin{figure*}[t]
\centering
\includegraphics[width=\linewidth]{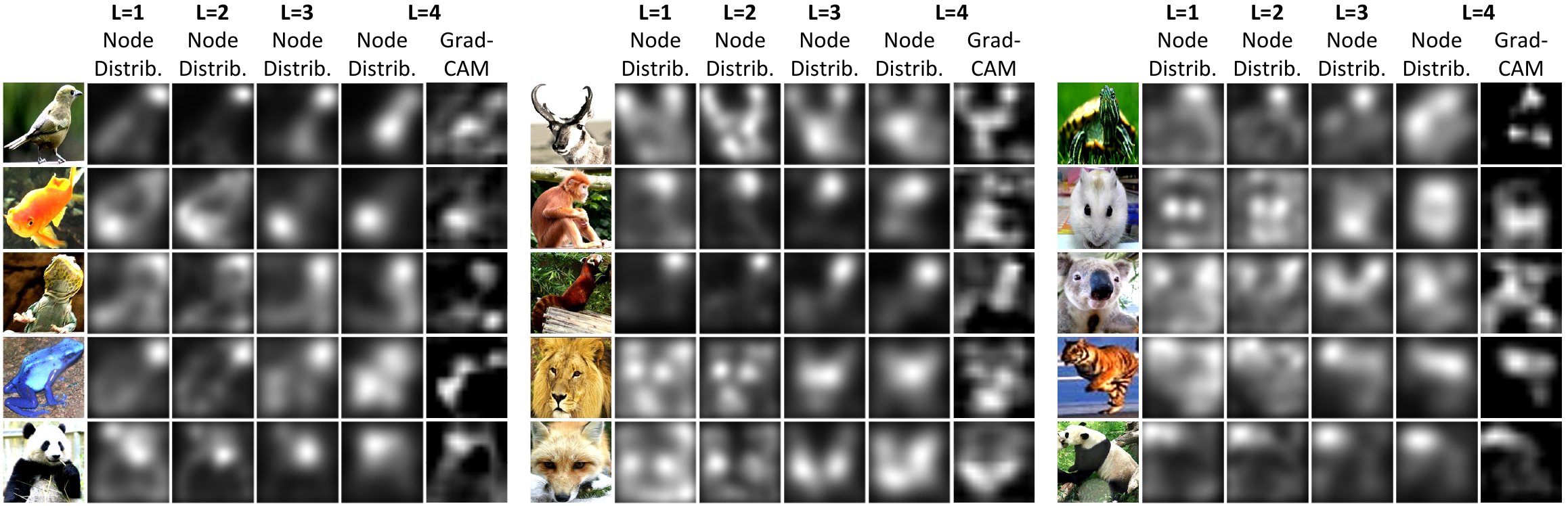}
\caption{Heatmaps of patterns. We use a heatmap to visualize the spatial distribution of the top-50\% patterns in the $L$-th layer of the explanatory graph with the highest inference scores. We also compare heatmaps with the grad-CAM~\cite{visualCNN_grad_2} of the feature map. Unlike the grad-CAM, our heatmaps mainly focus on the foreground of an object and uniformly pay attention to all parts, rather than only focus on most discriminative parts.}
\label{fig:heatmap}
\end{figure*}

We first trained/fine-tuned a CNN using object images of a category, which were cropped using object bounding boxes. Then, we set parameters {$\tau\!=\!0.1$}, {$M\!=\!15$}, {$T\!=\!20$}, and {$\beta\!=\!1$} to learn an explanatory graph for the CNN.\\
${\boldsymbol\bullet}$\textit{VGG-16:} The VGG-16 was first pre-trained using the 1.3M images in the ImageNet dataset~\cite{ImageNet}. We then fine-tuned all conv-layers of the VGG-16 using object images in a category. The loss for fine-tuning was for binary classification between the target category and background images. The VGG-16 has thirteen conv-layers and three fully connected layers. We selected the ninth, tenth, twelfth, and thirteenth conv-layers of the VGG-16 as four valid conv-layers, and accordingly, we built a four-layer graph. We extracted $N_{L,d}$ patterns from the $d$-th filter of the $L$-th layer, where we set $N_{L=1\,\textrm{or}\,2,d}=40$ and $N_{L=3\,\textrm{or}\,4,d}=20$.\\
${\boldsymbol\bullet}$\textit{Residual Networks:} Two residual networks, \emph{i.e.} the 50-layer and 152-layer ones, were used in experiments. The fine-tuning process for each network was exactly the same as that for VGG-16. We built a three-layer graph based on each residual network by selecting the last conv-layer with a $28\times28\times128$ feature output, the last conv-layer with a {$14\times14\times256$} feature map, and the last conv-layer with a {$7\times7\times512$} feature map as valid conv-layers. We set $N_{L=1,d}=40$, $N_{L=2,d}=20$, and $N_{L=3,d}=10$.\\
${\boldsymbol\bullet}$\textit{VAE-GAN:} For each category, we used the cropped object images to train a VAE-GAN. We learned a three-layer graph based on all three conv-layers of the encoder of the VAE-GAN. We set $N_{L=1,d}=52$, $N_{L=2,d}=26$, and $N_{L=3,d}=13$.

\subsection{Experiment 1: pattern visualization}

The global structure of an explanatory graph for a VGG-16 network is visualized in Fig.~\ref{fig:global}. We visualized detailed part patterns of graph nodes from the following three perspectives.

\begin{figure*}[t]
\centering
\includegraphics[width=\linewidth]{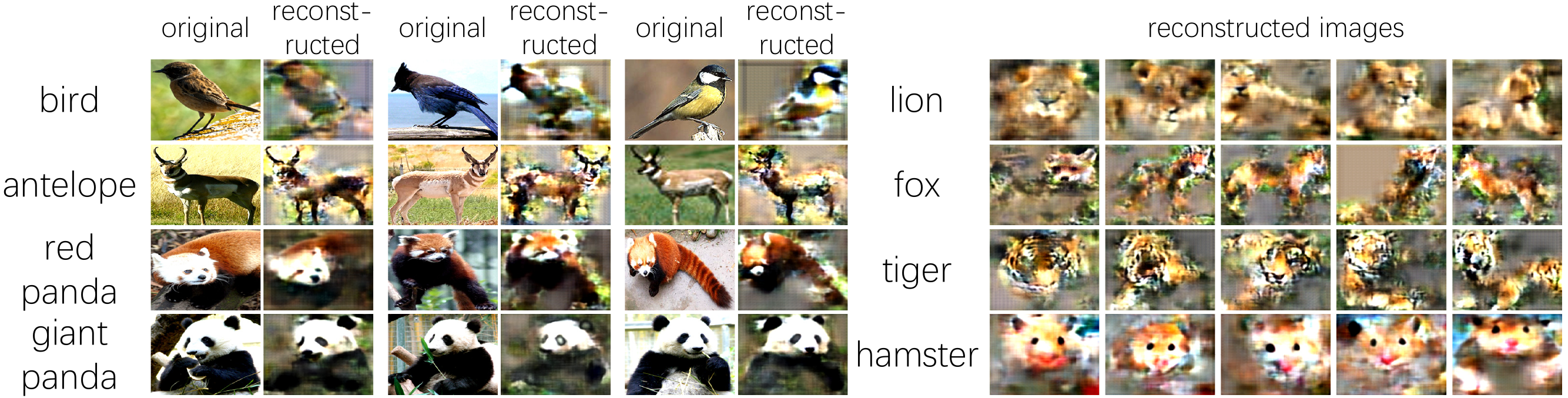}
\caption{Image synthesis result based on patterns activated on an image. The explanatory graph only encodes major part patterns hidden in conv-layers, rather than compress a CNN without information loss. Synthesis results demonstrate that the patterns are automatically learned to represent foreground appearance, and ignore background noises and trivial details of objects.}
\label{fig:reconstruct}
\end{figure*}

\textbf{Top-ranked patches:} For each image $I$, we performed the pattern inference on its feature maps. For a node $V$, we extracted a patch at the location of {${\bf p}_{\hat{x}_{V}}$}\footnote[4]{We projected the unit to the image to compute its position.} with a fixed scale of {$70\,pixels\!\times\!70\,pixels$} to represent $V$. Fig.~\ref{fig:patch} shows a pattern's image patches that had highest inference scores.

\textbf{Heatmaps of patterns:} Given inference results of patterns \emph{w.r.t.} a cropped object image $I$, we drew heatmaps to show the spatial distribution of the inferred patterns. We drew a heatmap for each layer $L$ of the graph. Each pattern {$V\in\Omega_{L}$} was visualized as a weighted Gaussian distribution $\alpha\cdot{\bf\mathcal N}(\mu={\bf p}_{V},\sigma_{V}^2)$\footnotemark[4] on the heatmap, where $\alpha=S_{V\rightarrow\hat{x}}^{I}$. Fig.~\ref{fig:heatmap} shows heatmaps of the top-50\% patterns with the highest scores of $S_{V\rightarrow\hat{x}}^{I}$.

\textbf{Pattern-based image synthesis:} We used the up-convolutional network~\cite{FeaVisual} to visualize part patterns of graph nodes. Given an object image $I$, we used the explanatory graph for pattern inference, \emph{i.e.} assigning each pattern $V$ with a certain neural unit {$\hat{x}_{V}$} as its position inference\footnotemark[4]. We considered the top-10\% patterns with highest scores of {$S_{V\rightarrow\hat{x}}^{I}$} as valid ones. We filtered out all neural responses of units, which were not assigned to valid patterns, from feature maps (setting these responses to zero). We selected the filtered feature map corresponding to the second graph layer and used the up-convolutional network to synthesize the filtered feature map to the input image. Fig.~\ref{fig:reconstruct} shows image-synthesis results, which can be regarded as the visualization of the inferred patterns.

\begin{figure*}[t]
\centering
\includegraphics[width=\linewidth]{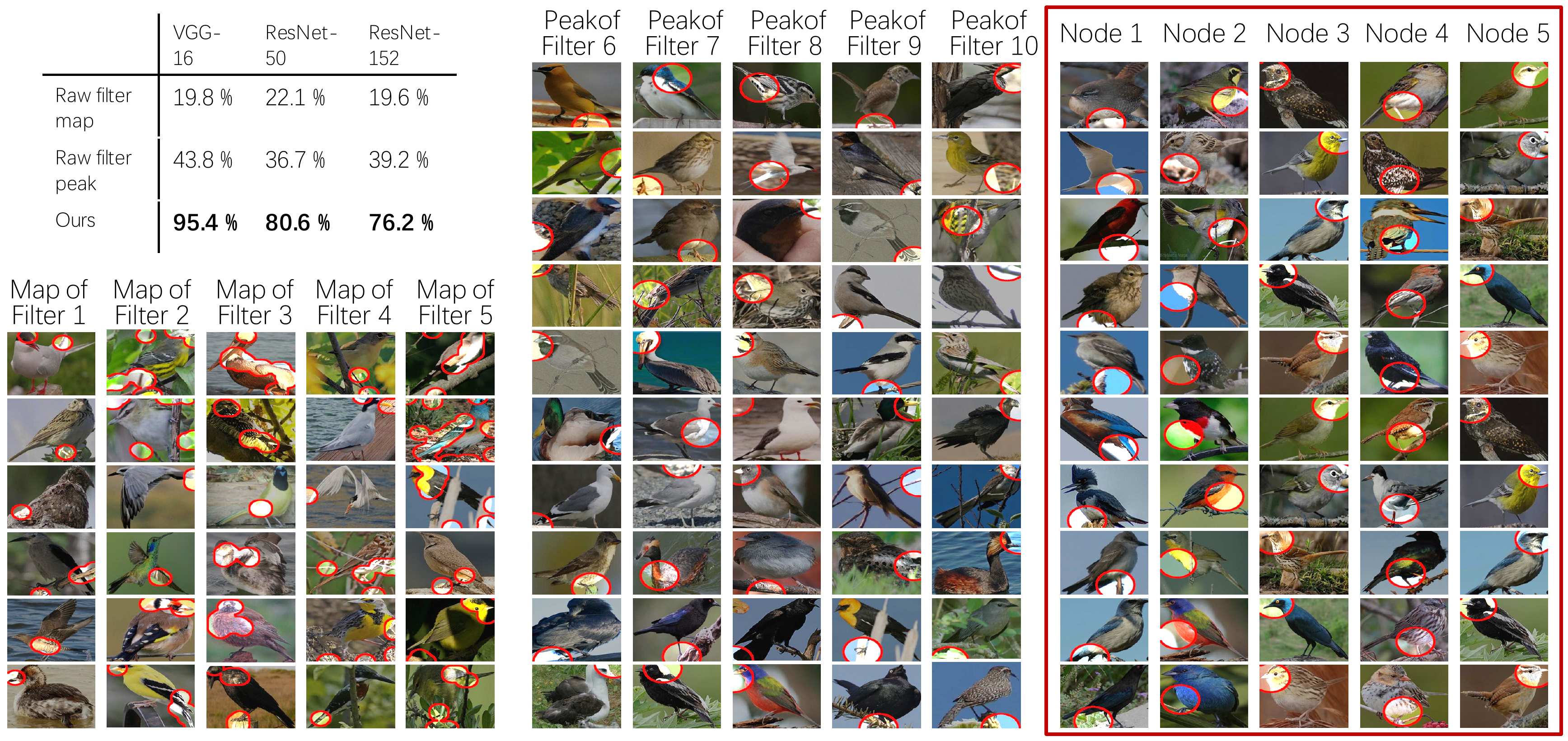}
\caption{Purity of part semantics (top-left). We compared patterns corresponding to nodes in the explanatory graph with patterns of raw filters. We draw raw feature maps of filters (left), the highest activation peaks on feature maps of filters (middle), and image regions corresponding to each node in the explanatory graph (right). Based on such visualization results, we use human users to annotate the semantic purity of each node/filter.}
\label{fig:interpretability}
\end{figure*}

\subsection{Experiment 2: semantic interpretability of patterns}

In this experiment, we evaluated whether or not each node pattern consistently represented the same object part through different images. Four explanatory graphs were built for a VGG-16 network, two residual networks, and a VAE-GAN. These networks were learned using the CUB200-2011 dataset~\cite{CUB200}. We used the following two metrics to measure the interpretability of node patterns.

\textbf{Part interpretability of patterns:} We mainly extracted patterns from high conv-layers, because as discussed in \cite{Interpretability}, high conv-layers contain large-scale part patterns. The evaluation metric was inspired by Zhou \emph{et al.}~\cite{CNNSemanticDeep}. For the pattern of a given node $V$, we used $V$ to make inferences among all images. We regarded inference results with the top-$K$ inference scores $S_{V}^{I_{i}}$ among all images as valid representations of $V$. We require the $K$ highest inference scores $S_{V}^{I_{i}}$ on images $\{I_1,\ldots,I_{k}\}$ to take about 30\% of the inference energy, \emph{i.e.} we use $\sum_{i=1}^{K}S_{V}^{I_{i}}=0.3\sum_{i\in{\bf I}}S_{V}^{I}$ to compute $K$. We asked human raters to count the number of inference results, which described the same object part, among the top $K$, in order to compute the purity of part semantics of pattern $V$.

The table in Fig.~\ref{fig:interpretability}(top-left) shows the semantic purity of the patterns in the second layer of the graph. Let the second graph layer correspond to the $L$-th conv-layer with $D$ filters. The \textit{raw filter maps} baseline used all neural activation in the feature map of a filter to describe a part. The \textit{raw filter peaks} baseline considered the highest peak on a filer's feature map as the part detection. Like our method, the two baselines also visualized top-$K'$ part inferences (the $K'$ feature maps' neural activations took 30\% of activation energies over all images). We back-propagated the center of the receptive field of each neural activation to the image plane and draw the image region corresponding to each neural activation. Fig.~\ref{fig:interpretability} compares the image region corresponding to each graph node and image regions corresponding to feature maps of each filter. Our graph nodes represented explicit object parts, but raw filters encoded mixed semantics.

\begin{figure}[t]
\centering
\includegraphics[width=0.9\linewidth]{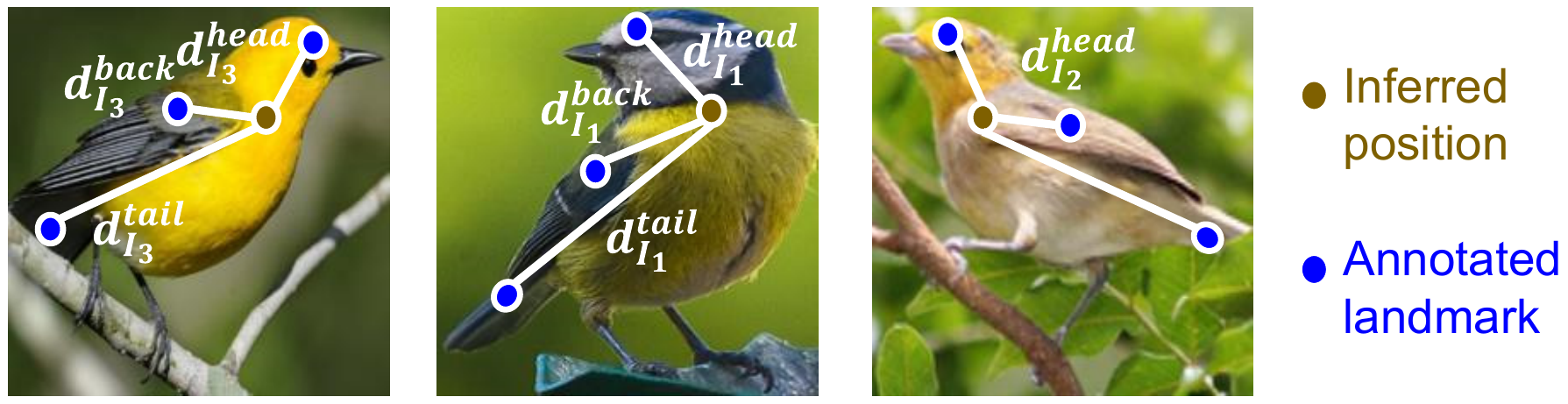}
\caption{Notation for the computation of location instability.}
\label{fig:instability}
\end{figure}

Because the baselines simply averaged the semantic purity among the $D$ filters, we also computed average semantic purities using the top-$D$ nodes with the highest scores of $\sum_{i\in{\bf I}}S_{V}^{I}$ to enable a fair comparison.

\begin{table}[t]
\centering
\caption{Location instability of patterns.}
\label{tab:stability}
\resizebox{1.0\linewidth}{!}{\begin{tabular}{l|cccc}
\hline
\!\!\!&\!\!\! {\small ResNet-50} \!\!\!&\!\!\! {\small ResNet-152} \!\!\!&\!\!\! VGG-16 \!\!\!& VAE-GAN\!\!\!\\
{\small Raw filter~\cite{CNNSemanticDeep}} & 0.1328 & 0.1346 & 0.1398 & 0.1944\\
Ours & {\bf0.0848} & {\bf0.0858} & {\bf0.0638} & {\bf0.1066}\\
\hline
{\cite{MiddleLevel}} & \multicolumn{4}{|c}{0.1341}\\
{\cite{CNNSemanticPart}} & \multicolumn{4}{|c}{0.2291}\\
\hline
\end{tabular}}
\end{table}

\begin{figure}[t]
\centering
\includegraphics[width=\linewidth]{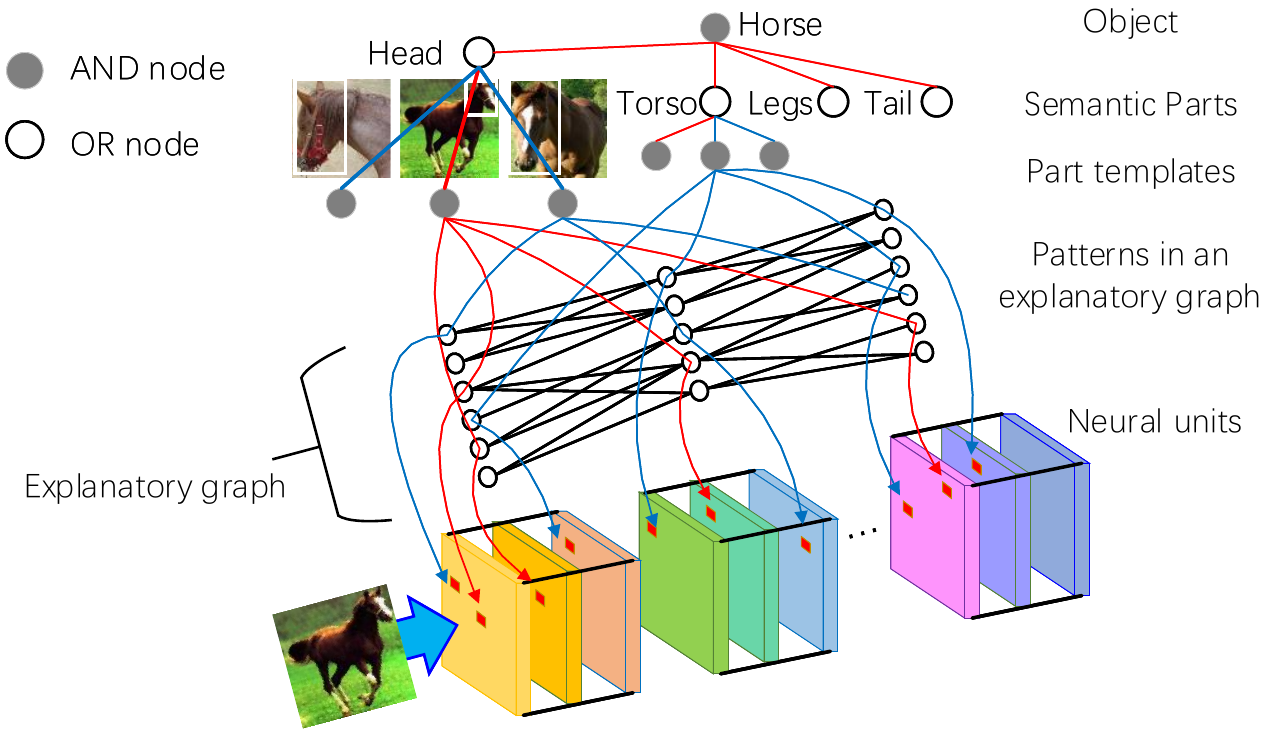}
\caption{Schematic illustration of an And-Or graph for semantic object parts. The AOG encodes a four-layer hierarchy for each semantic part, \emph{i.e.} the semantic part (OR node), part templates (AND node), latent part patterns (OR nodes, those from the explanatory graph), and neural units (terminal nodes). In the AOG, the OR node of semantic part contains a number of alternative appearance candidates as children. Each OR node of a latent part pattern encodes a list of neural units as alternative deformation candidates. Each AND node (\emph{e.g.} a part template) uses a number of latent part patterns to describe its compositional regions.}
\label{fig:hybrid}
\end{figure}

\textbf{Location instability of inference positions:} We defined the location instability for each pattern as another evaluation metric of pattern interpretability. We assumed that if a pattern was always triggered by the same object part through different images, then the distance between the pattern's inference position and a ground-truth landmark of the object part should not change a lot among various images.

As shown in Fig.~\ref{fig:instability}, given a testing image $I$, $d_{I}^{\textrm{head}}$, $d_{I}^{\textrm{back}}$, and $d_{I}^{\textrm{tail}}$ denote the distances between the inferred position of $V$ and ground-truth landmark positions of \textit{head}, \textit{back}, and \textit{tail} parts, respectively. These distances were normalized by the diagonal length of input images. Then, the node's location instability was given as $(\sqrt{var(d_{I}^{\textrm{head}})}+\sqrt{var(d_{I}^{\textrm{back}})}+\sqrt{var(d_{I}^{\textrm{tail}})})/3$, where $var(d_{I}^{\textrm{head}})$ denotes the variation of $d_{I}^{\textrm{head}}$ over different images.

We compared its location instability of an explanatory graph with three baselines. The first baseline treated each filter in a CNN as a detector of a certain pattern. Thus, given the feature map of a filter (after the ReLu operation), we used the method of \cite{CNNSemanticDeep} to localize the unit with the highest response value as the pattern position. The other two baselines were typical methods to extract middle-level features from images~\cite{MiddleLevel} and extract patterns from CNNs~\cite{CNNSemanticPart}, respectively. For each baseline, we chose the top-500 patterns, \emph{i.e.} 500 nodes with top scores in the explanatory graph, 500 filters with strongest activations in the CNN, and the top-500 middle-level features. For each pattern, we selected position inferences on the top-20 images with highest scores to compute the location instability. Table~\ref{tab:stability} compares the location instability of different baselines. Nodes in the explanatory graph had significantly lower location instability than patterns of baselines.

\subsection{Experiment 3: few-shot part localization}

\begin{table}[t]
\centering
\caption{Normalized distance of part localization on the CUB200-2011 dataset~\cite{CUB200}. The second column indicates whether the baseline used all object-box annotations in the category to fine-tune a CNN.}
\label{tab:CUB}
\resizebox{\linewidth}{!}{\begin{tabular}{c|lcc}
\hline
&\multicolumn{2}{r}{Method$\qquad$obj.-box fine-tune}&\\
\multirow{3}{*}{\large$\textrm{not learn}\atop\textrm{parts}$}
&{\small SS-DPM-Part~\cite{SSDPM}} & {N}
&{\small0.3469}
\\
&{\small PL-DPM-Part~\cite{PLDPM}} & {N}
&{\small0.3412}
\\
&{\small Part-Graph~\cite{SemanticPart}} & {N}
&{\small0.4889}
\\
\hline
\multirow{3}{*}{\large$\textrm{unsuper-learn\footnotemark[5]}\atop\textrm{parts}$}
&{\small CNN-PDD~\cite{CNNSemanticPart}} & {N}
&{\small0.2333}
\\
&{\small CNN-PDD-ft~\cite{CNNSemanticPart}} & {Y}
&{\small0.3269}
\\
&{\small\bf Ours} & {Y}
&{\small\bf 0.0862}
\\
\hline
\multirow{4}{*}{\large$\textrm{super-learn}\atop\textrm{parts}$}
&{\small fc7+linearSVM} & {Y}
&{\small0.3120}
\\
&{\small fc7+sp+linearSVM} & {Y}
&{\small0.3120}
\\
&{\small Fast-RCNN (1 ft)~\cite{FastRCNN}} & {N}
&{\small0.4517}
\\
&{\small Fast-RCNN (2 fts)~\cite{FastRCNN}} & {Y}
&{\small0.4131}
\\
\hline
\end{tabular}}
\end{table}

\begin{table*}[t]
\centering
\caption{Normalized distance of part localization on the VOC Part dataset~\cite{SemanticPart}. The second column indicates whether the baseline used all object-box annotations in the category to fine-tune a CNN.}
\label{tab:VOC}
\resizebox{0.9\linewidth}{!}{\begin{tabular}{c|l|c|ccccccc}
\hline
&\multicolumn{2}{r|}{$\qquad\qquad$obj.-box fine-tune}& bird & cat & cow & dog & horse & sheep & \textcolor{blue}{\bf Avg.}\\
\multirow{3}{*}{\large$\textrm{not learn}\atop\textrm{parts}$}
&{\small SS-DPM-Part~\cite{SSDPM}} \!&\! {N}
&{\small0.356}
&{\small0.270}
&{\small0.264}
&{\small0.242}
&{\small0.262}
&{\small0.286}
&\textcolor{blue}{\small0.280}
\\
&{\small PL-DPM-Part~\cite{PLDPM}} \!&\! {N}
&{\small0.294}
&{\small0.328}
&{\small0.282}
&{\small0.312}
&{\small0.321}
&{\small0.840}
&\textcolor{blue}{\small0.396}
\\
&{\small Part-Graph~\cite{SemanticPart}} \!&\! {N}
&{\small0.360}
&{\small0.208}
&{\small0.263}
&{\small0.205}
&{\small0.386}
&{\small0.500}
&\textcolor{blue}{\small0.320}
\\
\cline{1-3}
\multirow{3}{*}{\large$\textrm{unsuper-learn\footnotemark[5]}\atop\textrm{parts}$}
&{\small CNN-PDD~\cite{CNNSemanticPart}} \!&\! {N}
&{\small0.301}
&{\small0.246}
&{\small0.220}
&{\small0.248}
&{\small0.292}
&{\small0.254}
&\textcolor{blue}{\small0.260}
\\
&{\small CNN-PDD-ft~\cite{CNNSemanticPart}} \!&\! {Y}
&{\small0.358}
&{\small0.268}
&{\small0.220}
&{\small0.200}
&{\small0.302}
&{\small0.269}
&\textcolor{blue}{\small0.269}
\\
&{\small\bf Ours} \!&\! {Y}
&{\small\bf 0.162}
&{\small\bf 0.130}
&{\small0.258}
&{\small\bf 0.137}
&{\small\bf 0.181}
&{\small\bf 0.192}
&\textcolor{blue}{\small\bf 0.177}
\\
\cline{1-3}
\multirow{4}{*}{\large$\textrm{super-learn}\atop\textrm{parts}$}
&{\small fc7+linearSVM} \!&\! {Y}
&{\small0.247}
&{\small0.174}
&{\small0.251}
&{\small0.217}
&{\small0.261}
&{\small0.317}
&\textcolor{blue}{\small0.244}
\\
&{\small fc7+sp+linearSVM} \!&\! {Y}
&{\small0.247}
&{\small0.174}
&{\small\bf 0.249}
&{\small0.217}
&{\small0.261}
&{\small0.317}
&\textcolor{blue}{\small0.244}
\\
&{\small Fast-RCNN (1 ft)~\cite{FastRCNN}} \!&\! {N}
&{\small0.324}
&{\small0.324}
&{\small0.325}
&{\small0.272}
&{\small0.347}
&{\small0.314}
&\textcolor{blue}{\small0.318}
\\
&{\small Fast-RCNN (2 fts)~\cite{FastRCNN}} \!&\! {Y}
&{\small0.350}
&{\small0.295}
&{\small0.255}
&{\small0.293}
&{\small0.367}
&{\small0.260}
&\textcolor{blue}{\small0.303}
\\
\hline
\end{tabular}}
\end{table*}

\begin{table*}[t]
\centering
\caption{Accuracy of part localization evaluated by ``$IoU\geq0.5$'' on the Pascal VOC Part dataset~\cite{SemanticPart}. The second column indicates whether the baseline used all object annotations in the category to pre-finetune a CNN before learning the part.}
\label{tab:VOC_IOU}
\resizebox{0.8\linewidth}{!}{\begin{tabular}{c|l|c|ccccccc}
\hline
&\multicolumn{2}{r|}{$\qquad\qquad$obj.-box fine-tune}& bird & cat & cow & dog & horse & sheep & \textcolor{blue}{\bf Avg.}\\
\multirow{3}{*}{\large$\textrm{not learn}\atop\textrm{parts}$}
&{\small SS-DPM-Part~\cite{SSDPM}} \!&\! {N}
&{\small0.0}
&{\small1.3}
&{\small1.6}
&{\small1.9}
&{\small1.1}
&{\small3.3}
&\textcolor{blue}{\small1.5}
\\
&{\small PL-DPM-Part~\cite{PLDPM}} \!&\! {N}
&{\small0.5}
&{\small1.1}
&{\small4.4}
&{\small0.4}
&{\small0.0}
&{\small0.0}
&\textcolor{blue}{\small1.1}
\\
&{\small Part-Graph~\cite{SemanticPart}} \!&\! {N}
&{\small2.9}
&{\small22.6}
&{\small12.1}
&{\small11.0}
&{\small3.2}
&{\small0.0}
&\textcolor{blue}{\small8.6}
\\
\cline{1-3}
\multirow{1}{*}{\large$\textrm{unsuper-learn\footnotemark[5]}\atop\textrm{parts}$}
&{\small\bf Ours} \!&\! {Y}
&{\small\bf 20.2}
&{\small\bf 34.9}
&{\small8.2}
&{\small\bf 33.8}
&{\small10.0}
&{\small14.5}
&\textcolor{blue}{\small\bf 20.3}
\\
\cline{1-3}
\multirow{8}{*}{\large$\textrm{super-learn}\atop\textrm{parts}$}
&{\small fc7+linearSVM} \!&\! {Y}
&{\small8.0}
&{\small27.6}
&{\small7.1}
&{\small10.4}
&{\small16.1}
&{\small6.2}
&\textcolor{blue}{\small12.6}
\\
&{\small fc7+sp+linearSVM} \!&\! {Y}
&{\small8.0}
&{\small27.6}
&{\small7.1}
&{\small10.4}
&{\small\bf16.1}
&{\small6.2}
&\textcolor{blue}{\small12.6}
\\
&{\small fc7+RBF-SVM} \!&\! {Y}
&{\small5.3}
&{\small26.0}
&{\small7.7}
&{\small8.9}
&{\small14.7}
&{\small8.3}
&\textcolor{blue}{\small11.8}
\\
&{\small fc7+sp+RBF-SVM} \!&\! {Y}
&{\small5.0}
&{\small26.3}
&{\small7.1}
&{\small8.8}
&{\small15.1}
&{\small8.7}
&\textcolor{blue}{\small11.8}
\\
&{\small fc7+NN} \!&\! {Y}
&{\small1.9}
&{\small21.0}
&{\small3.8}
&{\small4.7}
&{\small3.6}
&{\small5.0}
&\textcolor{blue}{\small6.7}
\\
&{\small fc7+sp+NN} \!&\! {Y}
&{\small1.9}
&{\small21.0}
&{\small3.8}
&{\small4.7}
&{\small3.6}
&{\small5.0}
&\textcolor{blue}{\small6.7}
\\
&{\small Fast-RCNN (1 ft)~\cite{FastRCNN}} \!&\! {N}
&{\small2.1}
&{\small2.2}
&{\small2.2}
&{\small1.9}
&{\small1.4}
&{\small7.0}
&\textcolor{blue}{\small2.8}
\\
&{\small Fast-RCNN (2 fts)~\cite{FastRCNN}} \!&\! {Y}
&{\small7.7}
&{\small24.0}
&{\small\bf18.7}
&{\small18.0}
&{\small5.0}
&{\small\bf19.4}
&\textcolor{blue}{\small15.5}
\\
\hline
\end{tabular}}
\end{table*}

\begin{table*}[t]
\centering
\caption{Normalized distance of part localization on the ILSVRC 2013 DET Animal-Part dataset~\cite{CNNAoG}. The second column indicates whether the baseline used all object-box annotations in the category to fine-tune a CNN.}
\label{tab:imgnet}
\resizebox{\linewidth}{!}{\begin{tabular}{c|l|c|ccccccccccc}
\hline
&\multicolumn{2}{r|}{$\qquad\qquad$obj.-box fine-tune} &\!\! gold. \!\!&\!\! bird \!\!&\!\! frog \!\!&\!\! turt. \!\!&\!\! liza. \!\!&\!\! koala \!\!&\!\! lobs. \!\!&\!\! dog \!\!&\!\! fox \!\!&\!\! cat \!\!&\!\! lion\\
\multirow{3}{*}{\large$\textrm{not learn}\atop\textrm{parts}$}&
\!\!\! {\small SS-DPM-Part} \!\!\!&\!\!\! {N}
\!\!\!&\!\!\!{\small0.297}
\!\!\!&\!\!\!{\small0.280}
\!\!\!&\!\!\!{\small0.257}
\!\!\!&\!\!\!{\small0.255}
\!\!\!&\!\!\!{\small0.317}
\!\!\!&\!\!\!{\small0.222}
\!\!\!&\!\!\!{\small0.207}
\!\!\!&\!\!\!{\small0.239}
\!\!\!&\!\!\!{\small0.305}
\!\!\!&\!\!\!{\small0.308}
\!\!\!&\!\!\!{\small0.238}
\\
&\!\!\! {\small PL-DPM-Part} \!\!\!&\!\!\! {N}
\!\!\!&\!\!\!{\small0.273}
\!\!\!&\!\!\!{\small0.256}
\!\!\!&\!\!\!{\small0.271}
\!\!\!&\!\!\!{\small0.321}
\!\!\!&\!\!\!{\small0.327}
\!\!\!&\!\!\!{\small0.242}
\!\!\!&\!\!\!{\small0.194}
\!\!\!&\!\!\!{\small0.238}
\!\!\!&\!\!\!{\small0.619}
\!\!\!&\!\!\!{\small0.215}
\!\!\!&\!\!\!{\small0.239}
\\
&\!\!\! {\small Part-Graph} \!\!\!&\!\!\! {N}
\!\!\!&\!\!\!{\small0.363}
\!\!\!&\!\!\!{\small0.316}
\!\!\!&\!\!\!{\small0.241}
\!\!\!&\!\!\!{\small0.322}
\!\!\!&\!\!\!{\small0.419}
\!\!\!&\!\!\!{\small0.205}
\!\!\!&\!\!\!{\small0.218}
\!\!\!&\!\!\!{\small0.218}
\!\!\!&\!\!\!{\small0.343}
\!\!\!&\!\!\!{\small0.242}
\!\!\!&\!\!\!{\small0.162}
\\
\cline{1-3}
\multirow{3}{*}{\large$\textrm{unsuper-learn\footnotemark[5]}\atop\textrm{parts}$}
&\!\!\! {\small CNN-PDD} \!\!\!&\!\!\! {N}
\!\!\!&\!\!\!{\small0.316}
\!\!\!&\!\!\!{\small0.289}
\!\!\!&\!\!\!{\small0.229}
\!\!\!&\!\!\!{\small0.260}
\!\!\!&\!\!\!{\small0.335}
\!\!\!&\!\!\!{\small0.163}
\!\!\!&\!\!\!{\small0.190}
\!\!\!&\!\!\!{\small0.220}
\!\!\!&\!\!\!{\small0.212}
\!\!\!&\!\!\!{\small0.196}
\!\!\!&\!\!\!{\small0.174}
\\
&\!\!\! {\small CNN-PDD-ft} \!\!\!&\!\!\! {Y}
\!\!\!&\!\!\!{\small0.302}
\!\!\!&\!\!\!{\small0.236}
\!\!\!&\!\!\!{\small0.261}
\!\!\!&\!\!\!{\small0.231}
\!\!\!&\!\!\!{\small0.350}
\!\!\!&\!\!\!{\small0.168}
\!\!\!&\!\!\!{\small0.170}
\!\!\!&\!\!\!{\small0.177}
\!\!\!&\!\!\!{\small0.264}
\!\!\!&\!\!\!{\small0.270}
\!\!\!&\!\!\!{\small0.206}
\\
&\!\!\! {\small\bf Ours} \!\!\!&\!\!\! {Y}
\!\!\!&\!\!\!{\small\bf 0.090}
\!\!\!&\!\!\!{\small\bf 0.091}
\!\!\!&\!\!\!{\small\bf 0.095}
\!\!\!&\!\!\!{\small 0.167}
\!\!\!&\!\!\!{\small\bf 0.124}
\!\!\!&\!\!\!{\small\bf 0.084}
\!\!\!&\!\!\!{\small\bf 0.155}
\!\!\!&\!\!\!{\small 0.147}
\!\!\!&\!\!\!{\small\bf 0.081}
\!\!\!&\!\!\!{\small\bf 0.129}
\!\!\!&\!\!\!{\small\bf 0.074}
\\
\cline{1-3}
\multirow{4}{*}{\large$\textrm{super-learn}\atop\textrm{parts}$}
&\!\!\! {\small fc7+linearSVM} \!\!\!&\!\!\! {Y}
\!\!\!&\!\!\!{\small0.150}
\!\!\!&\!\!\!{\small0.318}
\!\!\!&\!\!\!{\small0.186}
\!\!\!&\!\!\!{\small0.150}
\!\!\!&\!\!\!{\small0.257}
\!\!\!&\!\!\!{\small0.156}
\!\!\!&\!\!\!{\small0.196}
\!\!\!&\!\!\!{\small0.136}
\!\!\!&\!\!\!{\small0.101}
\!\!\!&\!\!\!{\small0.138}
\!\!\!&\!\!\!{\small0.132}
\\
&\!\!\! {\small fc7+sp+linearSVM} \!\!\!&\!\!\! {Y}
\!\!\!&\!\!\!{\small0.150}
\!\!\!&\!\!\!{\small0.318}
\!\!\!&\!\!\!{\small0.186}
\!\!\!&\!\!\!{\small\bf 0.150}
\!\!\!&\!\!\!{\small0.254}
\!\!\!&\!\!\!{\small0.156}
\!\!\!&\!\!\!{\small0.196}
\!\!\!&\!\!\!{\small\bf 0.136}
\!\!\!&\!\!\!{\small0.101}
\!\!\!&\!\!\!{\small0.138}
\!\!\!&\!\!\!{\small0.132}
\\
&\!\!\! {\small Fast-RCNN (1 ft)} \!\!\!&\!\!\! {N}
\!\!\!&\!\!\!{\small0.261}
\!\!\!&\!\!\!{\small0.365}
\!\!\!&\!\!\!{\small0.265}
\!\!\!&\!\!\!{\small0.310}
\!\!\!&\!\!\!{\small0.353}
\!\!\!&\!\!\!{\small0.365}
\!\!\!&\!\!\!{\small0.289}
\!\!\!&\!\!\!{\small0.363}
\!\!\!&\!\!\!{\small0.255}
\!\!\!&\!\!\!{\small0.319}
\!\!\!&\!\!\!{\small0.251}
\\
&\!\!\! {\small Fast-RCNN (2 fts)} \!\!\!&\!\!\! {Y}
\!\!\!&\!\!\!{\small0.340}
\!\!\!&\!\!\!{\small0.351}
\!\!\!&\!\!\!{\small0.388}
\!\!\!&\!\!\!{\small0.327}
\!\!\!&\!\!\!{\small0.411}
\!\!\!&\!\!\!{\small0.119}
\!\!\!&\!\!\!{\small0.330}
\!\!\!&\!\!\!{\small0.368}
\!\!\!&\!\!\!{\small0.206}
\!\!\!&\!\!\!{\small0.170}
\!\!\!&\!\!\!{\small0.144}
\\
\hline
&\!\!\!&\!\!\!&\!\!\! tiger \!\!&\!\! bear \!\!&\!\! rabb. \!\!&\!\! hams. \!\!&\!\! squi. \!\!&\!\! horse \!\!\!&\!\!\! zebra \!\!\!&\!\!\! swine \!\!\!&\!\!\! hippo \!\!\!&\!\!\! catt. \!\!\!&\!\!\! sheep\\
\multirow{3}{*}{\large$\textrm{not learn}\atop\textrm{parts}$}&
\!\!\! {\small SS-DPM-Part} \!\!\!&\!\!\! {N}
\!\!\!&\!\!\!{\small0.144}
\!\!\!&\!\!\!{\small0.260}
\!\!\!&\!\!\!{\small0.272}
\!\!\!&\!\!\!{\small0.178}
\!\!\!&\!\!\!{\small0.261}
\!\!\!&\!\!\!{\small0.246}
\!\!\!&\!\!\!{\small\bf 0.206}
\!\!\!&\!\!\!{\small0.240}
\!\!\!&\!\!\!{\small0.234}
\!\!\!&\!\!\!{\small0.246}
\!\!\!&\!\!\!{\small0.205}
\\
&\!\!\! {\small PL-DPM-Part} \!\!\!&\!\!\! {N}
\!\!\!&\!\!\!{\small0.136}
\!\!\!&\!\!\!{\small0.323}
\!\!\!&\!\!\!{\small0.228}
\!\!\!&\!\!\!{\small0.186}
\!\!\!&\!\!\!{\small0.281}
\!\!\!&\!\!\!{\small0.322}
\!\!\!&\!\!\!{\small0.267}
\!\!\!&\!\!\!{\small0.297}
\!\!\!&\!\!\!{\small0.273}
\!\!\!&\!\!\!{\small0.271}
\!\!\!&\!\!\!{\small0.413}
\\
&\!\!\! {\small Part-Graph} \!\!\!&\!\!\! {N}
\!\!\!&\!\!\!{\small0.127}
\!\!\!&\!\!\!{\small0.224}
\!\!\!&\!\!\!{\small0.188}
\!\!\!&\!\!\!{\small0.131}
\!\!\!&\!\!\!{\small0.208}
\!\!\!&\!\!\!{\small0.296}
\!\!\!&\!\!\!{\small0.315}
\!\!\!&\!\!\!{\small0.306}
\!\!\!&\!\!\!{\small0.378}
\!\!\!&\!\!\!{\small0.333}
\!\!\!&\!\!\!{\small0.230}
\\
\cline{1-3}
\multirow{3}{*}{\large$\textrm{unsuper-learn\footnotemark[5]}\atop\textrm{parts}$}
&\!\!\! {\small CNN-PDD} \!\!\!&\!\!\! {N}
\!\!\!&\!\!\!{\small0.160}
\!\!\!&\!\!\!{\small0.223}
\!\!\!&\!\!\!{\small0.266}
\!\!\!&\!\!\!{\small0.156}
\!\!\!&\!\!\!{\small0.291}
\!\!\!&\!\!\!{\small0.261}
\!\!\!&\!\!\!{\small0.266}
\!\!\!&\!\!\!{\small\bf 0.189}
\!\!\!&\!\!\!{\small0.192}
\!\!\!&\!\!\!{\small0.201}
\!\!\!&\!\!\!{\small0.244}
\\
&\!\!\! {\small CNN-PDD-ft} \!\!\!&\!\!\! {Y}
\!\!\!&\!\!\!{\small0.256}
\!\!\!&\!\!\!{\small0.178}
\!\!\!&\!\!\!{\small0.167}
\!\!\!&\!\!\!{\small0.286}
\!\!\!&\!\!\!{\small0.237}
\!\!\!&\!\!\!{\small0.310}
\!\!\!&\!\!\!{\small0.321}
\!\!\!&\!\!\!{\small0.216}
\!\!\!&\!\!\!{\small0.257}
\!\!\!&\!\!\!{\small0.220}
\!\!\!&\!\!\!{\small0.179}
\\
&\!\!\! {\small\bf Ours} \!\!\!&\!\!\! {Y}
\!\!\!&\!\!\!{\small\bf 0.102}
\!\!\!&\!\!\!{\small\bf 0.121}
\!\!\!&\!\!\!{\small\bf 0.087}
\!\!\!&\!\!\!{\small\bf 0.097}
\!\!\!&\!\!\!{\small\bf 0.095}
\!\!\!&\!\!\!{\small\bf 0.189}
\!\!\!&\!\!\!{\small 0.212}
\!\!\!&\!\!\!{\small 0.212}
\!\!\!&\!\!\!{\small 0.151}
\!\!\!&\!\!\!{\small\bf 0.185}
\!\!\!&\!\!\!{\small\bf 0.124}
\\
\cline{1-3}
\multirow{4}{*}{\large$\textrm{super-learn}\atop\textrm{parts}$}
&\!\!\! {\small fc7+linearSVM} \!\!\!&\!\!\! {Y}
\!\!\!&\!\!\!{\small0.163}
\!\!\!&\!\!\!{\small0.122}
\!\!\!&\!\!\!{\small0.139}
\!\!\!&\!\!\!{\small0.110}
\!\!\!&\!\!\!{\small0.262}
\!\!\!&\!\!\!{\small0.205}
\!\!\!&\!\!\!{\small0.258}
\!\!\!&\!\!\!{\small0.201}
\!\!\!&\!\!\!{\small0.140}
\!\!\!&\!\!\!{\small0.256}
\!\!\!&\!\!\!{\small0.236}
\\
&\!\!\! {\small fc7+sp+linearSVM} \!\!\!&\!\!\! {Y}
\!\!\!&\!\!\!{\small0.163}
\!\!\!&\!\!\!{\small0.122}
\!\!\!&\!\!\!{\small0.139}
\!\!\!&\!\!\!{\small0.110}
\!\!\!&\!\!\!{\small0.262}
\!\!\!&\!\!\!{\small0.205}
\!\!\!&\!\!\!{\small0.258}
\!\!\!&\!\!\!{\small0.201}
\!\!\!&\!\!\!{\small\bf 0.140}
\!\!\!&\!\!\!{\small0.256}
\!\!\!&\!\!\!{\small0.236}
\\
&\!\!\! {\small Fast-RCNN (1 ft)} \!\!\!&\!\!\! {N}
\!\!\!&\!\!\!{\small0.260}
\!\!\!&\!\!\!{\small0.317}
\!\!\!&\!\!\!{\small0.255}
\!\!\!&\!\!\!{\small0.255}
\!\!\!&\!\!\!{\small0.169}
\!\!\!&\!\!\!{\small0.374}
\!\!\!&\!\!\!{\small0.322}
\!\!\!&\!\!\!{\small0.285}
\!\!\!&\!\!\!{\small0.265}
\!\!\!&\!\!\!{\small0.320}
\!\!\!&\!\!\!{\small0.277}
\\
&\!\!\! {\small Fast-RCNN (2 fts)} \!\!\!&\!\!\! {Y}
\!\!\!&\!\!\!{\small0.160}
\!\!\!&\!\!\!{\small0.230}
\!\!\!&\!\!\!{\small0.230}
\!\!\!&\!\!\!{\small0.178}
\!\!\!&\!\!\!{\small0.205}
\!\!\!&\!\!\!{\small0.346}
\!\!\!&\!\!\!{\small0.303}
\!\!\!&\!\!\!{\small0.212}
\!\!\!&\!\!\!{\small0.223}
\!\!\!&\!\!\!{\small0.228}
\!\!\!&\!\!\!{\small0.195}
\\
\hline
&\!\!\!&\!\!\!&\!\!\! ante. \!\!\!&\!\!\! camel \!\!\!&\!\!\! otter \!\!\!&\!\!\! arma. \!\!\!&\!\!\! monk. \!\!\!&\!\!\! elep. \!\!\!&\!\!\! red pa. \!\!\!&\!\!\! gia.pa. \!\!\!&\!\!\! \!\!\!&\!\!\! \!\!\!&\!\!\! \textcolor{blue}{\bf Avg.}\\
\multirow{3}{*}{\large$\textrm{not learn}\atop\textrm{parts}$}
&\!\!\! {\small SS-DPM-Part} \!\!\!&\!\!\! {N}
\!\!\!&\!\!\!{\small0.224}
\!\!\!&\!\!\!{\small0.277}
\!\!\!&\!\!\!{\small0.253}
\!\!\!&\!\!\!{\small0.283}
\!\!\!&\!\!\!{\small0.206}
\!\!\!&\!\!\!{\small0.219}
\!\!\!&\!\!\!{\small0.256}
\!\!\!&\!\!\!{\small0.129}
\!\!\!&\!\!\!
\!\!\!&\!\!\!
\!\!\!&\!\!\!{\small\textcolor{blue}{0.242}}
\\
&\!\!\! {\small PL-DPM-Part} \!\!\!&\!\!\! {N}
\!\!\!&\!\!\!{\small0.337}
\!\!\!&\!\!\!{\small0.261}
\!\!\!&\!\!\!{\small0.286}
\!\!\!&\!\!\!{\small0.295}
\!\!\!&\!\!\!{\small0.187}
\!\!\!&\!\!\!{\small0.264}
\!\!\!&\!\!\!{\small0.204}
\!\!\!&\!\!\!{\small0.505}
\!\!\!&\!\!\!
\!\!\!&\!\!\!
\!\!\!&\!\!\!{\small\textcolor{blue}{0.284}}
\\
&\!\!\! {\small Part-Graph} \!\!\!&\!\!\! {N}
\!\!\!&\!\!\!{\small0.216}
\!\!\!&\!\!\!{\small0.317}
\!\!\!&\!\!\!{\small0.227}
\!\!\!&\!\!\!{\small0.341}
\!\!\!&\!\!\!{\small0.159}
\!\!\!&\!\!\!{\small0.294}
\!\!\!&\!\!\!{\small0.276}
\!\!\!&\!\!\!{\small0.094}
\!\!\!&\!\!\!
\!\!\!&\!\!\!
\!\!\!&\!\!\!{\small\textcolor{blue}{0.257}}
\\
\cline{1-3}
\multirow{3}{*}{\large$\textrm{unsuper-learn\footnotemark[5]}\atop\textrm{parts}$}
&\!\!\! {\small CNN-PDD} \!\!\!&\!\!\! {N}
\!\!\!&\!\!\!{\small0.208}
\!\!\!&\!\!\!{\small0.193}
\!\!\!&\!\!\!{\small0.174}
\!\!\!&\!\!\!{\small0.299}
\!\!\!&\!\!\!{\small0.236}
\!\!\!&\!\!\!{\small0.214}
\!\!\!&\!\!\!{\small0.222}
\!\!\!&\!\!\!{\small0.179}
\!\!\!&\!\!\!
\!\!\!&\!\!\!
\!\!\!&\!\!\!{\small\textcolor{blue}{0.225}}
\\
&\!\!\! {\small CNN-PDD-ft} \!\!\!&\!\!\! {Y}
\!\!\!&\!\!\!{\small0.229}
\!\!\!&\!\!\!{\small0.253}
\!\!\!&\!\!\!{\small0.198}
\!\!\!&\!\!\!{\small0.308}
\!\!\!&\!\!\!{\small0.273}
\!\!\!&\!\!\!{\small0.189}
\!\!\!&\!\!\!{\small0.208}
\!\!\!&\!\!\!{\small0.275}
\!\!\!&\!\!\!
\!\!\!&\!\!\!
\!\!\!&\!\!\!{\small\textcolor{blue}{0.240}}
\\
&\!\!\! {\small\bf Ours} \!\!\!&\!\!\! {Y}
\!\!\!&\!\!\!{\small\bf 0.093}
\!\!\!&\!\!\!{\small\bf 0.120}
\!\!\!&\!\!\!{\small\bf 0.102}
\!\!\!&\!\!\!{\small\bf 0.188}
\!\!\!&\!\!\!{\small\bf 0.086}
\!\!\!&\!\!\!{\small 0.174}
\!\!\!&\!\!\!{\small\bf 0.104}
\!\!\!&\!\!\!{\small\bf 0.073}
\!\!\!&\!\!\!
\!\!\!&\!\!\!
\!\!\!&\!\!\!{\small\textcolor{blue}{\bf 0.125}}
\\
\cline{1-3}
\multirow{4}{*}{\large$\textrm{super-learn}\atop\textrm{parts}$}
&\!\!\! {\small fc7+linearSVM} \!\!\!&\!\!\! {Y}
\!\!\!&\!\!\!{\small0.164}
\!\!\!&\!\!\!{\small0.190}
\!\!\!&\!\!\!{\small0.140}
\!\!\!&\!\!\!{\small0.252}
\!\!\!&\!\!\!{\small0.256}
\!\!\!&\!\!\!{\small0.176}
\!\!\!&\!\!\!{\small0.215}
\!\!\!&\!\!\!{\small0.116}
\!\!\!&\!\!\!
\!\!\!&\!\!\!
\!\!\!&\!\!\!{\small\textcolor{blue}{0.184}}
\\
&\!\!\! {\small fc7+sp+linearSVM} \!\!\!&\!\!\! {Y}
\!\!\!&\!\!\!{\small0.164}
\!\!\!&\!\!\!{\small0.190}
\!\!\!&\!\!\!{\small0.140}
\!\!\!&\!\!\!{\small0.250}
\!\!\!&\!\!\!{\small0.256}
\!\!\!&\!\!\!{\small0.176}
\!\!\!&\!\!\!{\small0.215}
\!\!\!&\!\!\!{\small0.116}
\!\!\!&\!\!\!
\!\!\!&\!\!\!
\!\!\!&\!\!\!{\small\textcolor{blue}{0.184}}
\\
&\!\!\! {\small Fast-RCNN (1 ft)} \!\!\!&\!\!\! {N}
\!\!\!&\!\!\!{\small0.255}
\!\!\!&\!\!\!{\small0.351}
\!\!\!&\!\!\!{\small0.340}
\!\!\!&\!\!\!{\small0.324}
\!\!\!&\!\!\!{\small0.334}
\!\!\!&\!\!\!{\small0.256}
\!\!\!&\!\!\!{\small0.336}
\!\!\!&\!\!\!{\small0.274}
\!\!\!&\!\!\!
\!\!\!&\!\!\!
\!\!\!&\!\!\!{\small\textcolor{blue}{0.299}}
\\
&\!\!\! {\small Fast-RCNN (2 fts)} \!\!\!&\!\!\! {Y}
\!\!\!&\!\!\!{\small0.175}
\!\!\!&\!\!\!{\small0.247}
\!\!\!&\!\!\!{\small0.280}
\!\!\!&\!\!\!{\small0.319}
\!\!\!&\!\!\!{\small0.193}
\!\!\!&\!\!\!{\small\bf 0.125}
\!\!\!&\!\!\!{\small0.213}
\!\!\!&\!\!\!{\small0.160}
\!\!\!&\!\!\!
\!\!\!&\!\!\!
\!\!\!&\!\!\!{\small\textcolor{blue}{0.246}}
\\
\hline
\end{tabular}}
\end{table*}

\begin{table*}[h]
\centering
\caption{Accuracy of part localization evaluated by ``$IoU\geq0.5$'' on the ILSVRC 2013 DET Animal-Part dataset~\cite{CNNAoG}. The second column indicates whether the baseline used all object annotations in the category to pre-finetune a CNN before learning the part.}
\label{tab:imgnet_IOU}
\resizebox{0.99\linewidth}{!}{\begin{tabular}{l|c|cccccccccccccccc}
\hline
\multicolumn{2}{r|}{$\qquad\qquad$obj.-box finetune} &\!\!\! gold. \!\!\!\!&\!\!\!\! bird \!\!\!\!&\!\!\!\! frog \!\!\!\!&\!\!\!\! turt. \!\!\!\!&\!\!\!\! liza. \!\!\!\!&\!\!\!\! koala \!\!\!\!&\!\!\!\! lobs. \!\!\!\!&\!\!\!\! dog \!\!\!\!&\!\!\!\! fox \!\!\!\!&\!\!\!\! cat \!\!\!\!&\!\!\!\! lion \!\!\!\!&\!\!\!\! tiger \!\!\!\!&\!\!\!\! bear \!\!\!\!&\!\!\!\! rabb. \!\!\!\!&\!\!\!\! hams. \!\!\!\!&\!\!\!\! squi.\\
\!\!\! {\small SS-DPM-Part~\cite{SSDPM}} \!\!\!\!\!&\!\!\!\!\! {N}
\!\!\!\!\!&\!\!\!\!\!1.5
\!\!\!\!\!&\!\!\!\!\!0.0
\!\!\!\!\!&\!\!\!\!\!1.2
\!\!\!\!\!&\!\!\!\!\!2.6
\!\!\!\!\!&\!\!\!\!\!0.7
\!\!\!\!\!&\!\!\!\!\!8.8
\!\!\!\!\!&\!\!\!\!\!1.4
\!\!\!\!\!&\!\!\!\!\!5.2
\!\!\!\!\!&\!\!\!\!\!0.0
\!\!\!\!\!&\!\!\!\!\!10.9
\!\!\!\!\!&\!\!\!\!\!13.4
\!\!\!\!\!&\!\!\!\!\!20.4
\!\!\!\!\!&\!\!\!\!\!7.0
\!\!\!\!\!&\!\!\!\!\!0.5
\!\!\!\!\!&\!\!\!\!\!6.5
\!\!\!\!\!&\!\!\!\!\!0.5
\\
\!\!\! {\small PL-DPM-Part~\cite{PLDPM}} \!\!\!\!\!&\!\!\!\!\! {N}
\!\!\!\!\!&\!\!\!\!\!0.0
\!\!\!\!\!&\!\!\!\!\!1.0
\!\!\!\!\!&\!\!\!\!\!0.0
\!\!\!\!\!&\!\!\!\!\!0.6
\!\!\!\!\!&\!\!\!\!\!0.0
\!\!\!\!\!&\!\!\!\!\!3.3
\!\!\!\!\!&\!\!\!\!\!0.0
\!\!\!\!\!&\!\!\!\!\!3.3
\!\!\!\!\!&\!\!\!\!\!0.0
\!\!\!\!\!&\!\!\!\!\!23.8
\!\!\!\!\!&\!\!\!\!\!8.8
\!\!\!\!\!&\!\!\!\!\!3.6
\!\!\!\!\!&\!\!\!\!\!0.0
\!\!\!\!\!&\!\!\!\!\!1.6
\!\!\!\!\!&\!\!\!\!\!22.3
\!\!\!\!\!&\!\!\!\!\!0.0
\\
\!\!\! {\small Part-Graph~\cite{SemanticPart}} \!\!\!\!\!&\!\!\!\!\! {N}
\!\!\!\!\!&\!\!\!\!\!2.0
\!\!\!\!\!&\!\!\!\!\!5.5
\!\!\!\!\!&\!\!\!\!\!5.9
\!\!\!\!\!&\!\!\!\!\!6.5
\!\!\!\!\!&\!\!\!\!\!7.4
\!\!\!\!\!&\!\!\!\!\!12.1
\!\!\!\!\!&\!\!\!\!\!3.5
\!\!\!\!\!&\!\!\!\!\!9.0
\!\!\!\!\!&\!\!\!\!\!1.9
\!\!\!\!\!&\!\!\!\!\!18.7
\!\!\!\!\!&\!\!\!\!\!40.7
\!\!\!\!\!&\!\!\!\!\!56.1
\!\!\!\!\!&\!\!\!\!\!15.0
\!\!\!\!\!&\!\!\!\!\!27.3
\!\!\!\!\!&\!\!\!\!\!37.7
\!\!\!\!\!&\!\!\!\!\!21.4
\\
\!\!\! {\small fc7+linearSVM} \!\!\!\!\!&\!\!\!\!\! {Y}
\!\!\!\!\!&\!\!\!\!\!20.0
\!\!\!\!\!&\!\!\!\!\!2.0
\!\!\!\!\!&\!\!\!\!\!13.5
\!\!\!\!\!&\!\!\!\!\!20.8
\!\!\!\!\!&\!\!\!\!\!7.4
\!\!\!\!\!&\!\!\!\!\!30.2
\!\!\!\!\!&\!\!\!\!\!1.4
\!\!\!\!\!&\!\!\!\!\!27.5
\!\!\!\!\!&\!\!\!\!\!55.9
\!\!\!\!\!&\!\!\!\!\!39.4
\!\!\!\!\!&\!\!\!\!\!43.3
\!\!\!\!\!&\!\!\!\!\!27.0
\!\!\!\!\!&\!\!\!\!\!46.5
\!\!\!\!\!&\!\!\!\!\!44.3
\!\!\!\!\!&\!\!\!\!\!60.5
\!\!\!\!\!&\!\!\!\!\!8.8
\\
\!\!\! {\small fc7+RBF-SVM} \!\!\!\!\!&\!\!\!\!\! {Y}
\!\!\!\!\!&\!\!\!\!\!4.5
\!\!\!\!\!&\!\!\!\!\!0.0
\!\!\!\!\!&\!\!\!\!\!2.4
\!\!\!\!\!&\!\!\!\!\!24.7
\!\!\!\!\!&\!\!\!\!\!5.9
\!\!\!\!\!&\!\!\!\!\!34.0
\!\!\!\!\!&\!\!\!\!\!0.7
\!\!\!\!\!&\!\!\!\!\!15.6
\!\!\!\!\!&\!\!\!\!\!29.9
\!\!\!\!\!&\!\!\!\!\!42.5
\!\!\!\!\!&\!\!\!\!\!53.1
\!\!\!\!\!&\!\!\!\!\!39.3
\!\!\!\!\!&\!\!\!\!\!19.0
\!\!\!\!\!&\!\!\!\!\!44.8
\!\!\!\!\!&\!\!\!\!\!41.4
\!\!\!\!\!&\!\!\!\!\!0.9
\\
\!\!\! {\small fc7+NN} \!\!\!\!\!&\!\!\!\!\! {Y}
\!\!\!\!\!&\!\!\!\!\!1.0
\!\!\!\!\!&\!\!\!\!\!0.0
\!\!\!\!\!&\!\!\!\!\!1.2
\!\!\!\!\!&\!\!\!\!\!7.1
\!\!\!\!\!&\!\!\!\!\!2.2
\!\!\!\!\!&\!\!\!\!\!28.4
\!\!\!\!\!&\!\!\!\!\!1.4
\!\!\!\!\!&\!\!\!\!\!5.2
\!\!\!\!\!&\!\!\!\!\!19.4
\!\!\!\!\!&\!\!\!\!\!20.2
\!\!\!\!\!&\!\!\!\!\!52.1
\!\!\!\!\!&\!\!\!\!\!39.8
\!\!\!\!\!&\!\!\!\!\!5.0
\!\!\!\!\!&\!\!\!\!\!17.5
\!\!\!\!\!&\!\!\!\!\!32.6
\!\!\!\!\!&\!\!\!\!\!0.5
\\
\!\!\! {\small fc7+sp+linearSVM} \!\!\!\!\!&\!\!\!\!\! {Y}
\!\!\!\!\!&\!\!\!\!\!20.0
\!\!\!\!\!&\!\!\!\!\!2.0
\!\!\!\!\!&\!\!\!\!\!13.5
\!\!\!\!\!&\!\!\!\!\!20.8
\!\!\!\!\!&\!\!\!\!\!7.4
\!\!\!\!\!&\!\!\!\!\!30.2
\!\!\!\!\!&\!\!\!\!\!1.4
\!\!\!\!\!&\!\!\!\!\!27.5
\!\!\!\!\!&\!\!\!\!\!55.9
\!\!\!\!\!&\!\!\!\!\!39.4
\!\!\!\!\!&\!\!\!\!\!43.3
\!\!\!\!\!&\!\!\!\!\!27.0
\!\!\!\!\!&\!\!\!\!\!{\bf46.5}
\!\!\!\!\!&\!\!\!\!\!44.3
\!\!\!\!\!&\!\!\!\!\!60.5
\!\!\!\!\!&\!\!\!\!\!8.8
\\
\!\!\! {\small fc7+sp+RBF-SVM} \!\!\!\!\!&\!\!\!\!\! {Y}
\!\!\!\!\!&\!\!\!\!\!4.5
\!\!\!\!\!&\!\!\!\!\!0.0
\!\!\!\!\!&\!\!\!\!\!1.8
\!\!\!\!\!&\!\!\!\!\!{\bf24.7}
\!\!\!\!\!&\!\!\!\!\!4.4
\!\!\!\!\!&\!\!\!\!\!34.4
\!\!\!\!\!&\!\!\!\!\!0.7
\!\!\!\!\!&\!\!\!\!\!14.7
\!\!\!\!\!&\!\!\!\!\!29.9
\!\!\!\!\!&\!\!\!\!\!41.5
\!\!\!\!\!&\!\!\!\!\!53.1
\!\!\!\!\!&\!\!\!\!\!38.8
\!\!\!\!\!&\!\!\!\!\!19.0
\!\!\!\!\!&\!\!\!\!\!44.3
\!\!\!\!\!&\!\!\!\!\!41.9
\!\!\!\!\!&\!\!\!\!\!0.9
\\
\!\!\! {\small fc7+sp+NN} \!\!\!\!\!&\!\!\!\!\! {Y}
\!\!\!\!\!&\!\!\!\!\!1.0
\!\!\!\!\!&\!\!\!\!\!0.0
\!\!\!\!\!&\!\!\!\!\!1.2
\!\!\!\!\!&\!\!\!\!\!7.1
\!\!\!\!\!&\!\!\!\!\!2.2
\!\!\!\!\!&\!\!\!\!\!28.4
\!\!\!\!\!&\!\!\!\!\!1.4
\!\!\!\!\!&\!\!\!\!\!5.2
\!\!\!\!\!&\!\!\!\!\!19.4
\!\!\!\!\!&\!\!\!\!\!20.2
\!\!\!\!\!&\!\!\!\!\!52.1
\!\!\!\!\!&\!\!\!\!\!39.8
\!\!\!\!\!&\!\!\!\!\!5.0
\!\!\!\!\!&\!\!\!\!\!17.5
\!\!\!\!\!&\!\!\!\!\!32.6
\!\!\!\!\!&\!\!\!\!\!0.5
\\
\!\!\! {\small Fast-RCNN (1 ft)~\cite{FastRCNN}} \!\!\!\!\!&\!\!\!\!\! {N}
\!\!\!\!\!&\!\!\!\!\!5.0
\!\!\!\!\!&\!\!\!\!\!0.5
\!\!\!\!\!&\!\!\!\!\!1.8
\!\!\!\!\!&\!\!\!\!\!2.6
\!\!\!\!\!&\!\!\!\!\!3.7
\!\!\!\!\!&\!\!\!\!\!3.3
\!\!\!\!\!&\!\!\!\!\!0
\!\!\!\!\!&\!\!\!\!\!0.5
\!\!\!\!\!&\!\!\!\!\!28.9
\!\!\!\!\!&\!\!\!\!\!11.4
\!\!\!\!\!&\!\!\!\!\!22.2
\!\!\!\!\!&\!\!\!\!\!11.7
\!\!\!\!\!&\!\!\!\!\!2.5
\!\!\!\!\!&\!\!\!\!\!20.2
\!\!\!\!\!&\!\!\!\!\!27.9
\!\!\!\!\!&\!\!\!\!\!36.3
\\
\!\!\! {\small Fast-RCNN (2 fts)~\cite{FastRCNN}} \!\!\!\!\!&\!\!\!\!\! {Y}
\!\!\!\!\!&\!\!\!\!\!4.5
\!\!\!\!\!&\!\!\!\!\!5.0
\!\!\!\!\!&\!\!\!\!\!2.4
\!\!\!\!\!&\!\!\!\!\!4.5
\!\!\!\!\!&\!\!\!\!\!2.2
\!\!\!\!\!&\!\!\!\!\!68.8
\!\!\!\!\!&\!\!\!\!\!1.4
\!\!\!\!\!&\!\!\!\!\!9.0
\!\!\!\!\!&\!\!\!\!\!46.0
\!\!\!\!\!&\!\!\!\!\!{\bf50.8}
\!\!\!\!\!&\!\!\!\!\!61.3
\!\!\!\!\!&\!\!\!\!\!{\bf65.8}
\!\!\!\!\!&\!\!\!\!\!29.0
\!\!\!\!\!&\!\!\!\!\!30.1
\!\!\!\!\!&\!\!\!\!\!56.3
\!\!\!\!\!&\!\!\!\!\!{\bf40.9}
\\
\!\!\! {\small Ours} \!\!\!\!\!&\!\!\!\!\! {Y}
\!\!\!\!\!&\!\!\!\!\!{\bf 33.0}
\!\!\!\!\!&\!\!\!\!\!{\bf 40.3}
\!\!\!\!\!&\!\!\!\!\!{\bf 48.8}
\!\!\!\!\!&\!\!\!\!\!18.2
\!\!\!\!\!&\!\!\!\!\!{\bf 21.4}
\!\!\!\!\!&\!\!\!\!\!{\bf 61.9}
\!\!\!\!\!&\!\!\!\!\!{\bf 3.5}
\!\!\!\!\!&\!\!\!\!\!{\bf 30.3}
\!\!\!\!\!&\!\!\!\!\!{\bf 62.1}
\!\!\!\!\!&\!\!\!\!\!26.4
\!\!\!\!\!&\!\!\!\!\!{\bf 61.9}
\!\!\!\!\!&\!\!\!\!\!49.5
\!\!\!\!\!&\!\!\!\!\!36.0
\!\!\!\!\!&\!\!\!\!\!{\bf 65.6}
\!\!\!\!\!&\!\!\!\!\!{\bf 64.7}
\!\!\!\!\!&\!\!\!\!\!25.6
\\
\hline
\!\!\!\!\!&\!\!\!\!\! \!\!\!\!\!&\!\!\!\!\! horse \!\!\!\!\!&\!\!\!\!\! zebra \!\!\!\!\!&\!\!\!\!\! swine \!\!\!\!\!&\!\!\!\!\! hippo \!\!\!\!\!&\!\!\!\!\! catt. \!\!\!\!\!&\!\!\!\!\! sheep \!\!\!\!\!&\!\!\!\!\! ante. \!\!\!\!\!&\!\!\!\!\! camel \!\!\!\!\!&\!\!\!\!\! otter \!\!\!\!\!&\!\!\!\!\! arma. \!\!\!\!\!&\!\!\!\!\! monk. \!\!\!\!\!&\!\!\!\!\! elep. \!\!\!\!\!&\!\!\!\!\! red pa. \!\!\!\!\!&\!\!\!\!\! gia.pa. \!\!\!\!\!&\!\!\!\!\! \!\!\!\!\!&\!\!\!\!\! \textcolor{blue}{\bf Avg.}\\
\!\!\! {\small SS-DPM-Part~\cite{SSDPM}} \!\!\!\!\!&\!\!\!\!\! {N}
\!\!\!\!\!&\!\!\!\!\!9.5
\!\!\!\!\!&\!\!\!\!\!1.1
\!\!\!\!\!&\!\!\!\!\!0.6
\!\!\!\!\!&\!\!\!\!\!1.1
\!\!\!\!\!&\!\!\!\!\!7.0
\!\!\!\!\!&\!\!\!\!\!14.7
\!\!\!\!\!&\!\!\!\!\!12.4
\!\!\!\!\!&\!\!\!\!\!0.9
\!\!\!\!\!&\!\!\!\!\!0.5
\!\!\!\!\!&\!\!\!\!\!4.5
\!\!\!\!\!&\!\!\!\!\!12.4
\!\!\!\!\!&\!\!\!\!\!11.8
\!\!\!\!\!&\!\!\!\!\!2.2
\!\!\!\!\!&\!\!\!\!\!49.1
\!\!\!\!\!&\!\!\!\!\!
\!\!\!&\!\!
\!\textcolor{blue}{7.0}
\\
\!\!\! {\small PL-DPM-Part~\cite{PLDPM}} \!\!\!\!\!&\!\!\!\!\! {N}
\!\!\!\!\!&\!\!\!\!\!5.8
\!\!\!\!\!&\!\!\!\!\!0.0
\!\!\!\!\!&\!\!\!\!\!0.6
\!\!\!\!\!&\!\!\!\!\!0.5
\!\!\!\!\!&\!\!\!\!\!0.5
\!\!\!\!\!&\!\!\!\!\!0.0
\!\!\!\!\!&\!\!\!\!\!0.0
\!\!\!\!\!&\!\!\!\!\!0.0
\!\!\!\!\!&\!\!\!\!\!0.0
\!\!\!\!\!&\!\!\!\!\!0.0
\!\!\!\!\!&\!\!\!\!\!9.1
\!\!\!\!\!&\!\!\!\!\!2.6
\!\!\!\!\!&\!\!\!\!\!28.1
\!\!\!\!\!&\!\!\!\!\!0.0
\!\!\!\!\!&\!\!\!\!\!
\!\!\!\!\!&\!\!\!\!\!\textcolor{blue}{3.9}
\\
\!\!\! {\small Part-Graph~\cite{SemanticPart}} \!\!\!\!\!&\!\!\!\!\! {N}
\!\!\!\!\!&\!\!\!\!\!10.0
\!\!\!\!\!&\!\!\!\!\!13.0
\!\!\!\!\!&\!\!\!\!\!4.9
\!\!\!\!\!&\!\!\!\!\!4.3
\!\!\!\!\!&\!\!\!\!\!7.0
\!\!\!\!\!&\!\!\!\!\!19.0
\!\!\!\!\!&\!\!\!\!\!23.0
\!\!\!\!\!&\!\!\!\!\!5.6
\!\!\!\!\!&\!\!\!\!\!18.2
\!\!\!\!\!&\!\!\!\!\!6.6
\!\!\!\!\!&\!\!\!\!\!18.3
\!\!\!\!\!&\!\!\!\!\!2.6
\!\!\!\!\!&\!\!\!\!\!16.2
\!\!\!\!\!&\!\!\!\!\!58.6
\!\!\!\!\!&\!\!\!\!\!
\!\!\!\!\!&\!\!\!\!\!\textcolor{blue}{15.9}
\\
\!\!\! {\small fc7+linearSVM} \!\!\!\!\!&\!\!\!\!\! {Y}
\!\!\!\!\!&\!\!\!\!\!16.3
\!\!\!\!\!&\!\!\!\!\!10.7
\!\!\!\!\!&\!\!\!\!\!22.0
\!\!\!\!\!&\!\!\!\!\!31.9
\!\!\!\!\!&\!\!\!\!\!4.9
\!\!\!\!\!&\!\!\!\!\!20.2
\!\!\!\!\!&\!\!\!\!\!26.3
\!\!\!\!\!&\!\!\!\!\!23.7
\!\!\!\!\!&\!\!\!\!\!35.3
\!\!\!\!\!&\!\!\!\!\!11.6
\!\!\!\!\!&\!\!\!\!\!12.4
\!\!\!\!\!&\!\!\!\!\!36.8
\!\!\!\!\!&\!\!\!\!\!22.8
\!\!\!\!\!&\!\!\!\!\!48.6
\!\!\!\!\!&\!\!\!\!\!
\!\!\!\!\!&\!\!\!\!\!\textcolor{blue}{25.7}
\\
\!\!\! {\small fc7+RBF-SVM} \!\!\!\!\!&\!\!\!\!\! {Y}
\!\!\!\!\!&\!\!\!\!\!7.9
\!\!\!\!\!&\!\!\!\!\!27.1
\!\!\!\!\!&\!\!\!\!\!7.3
\!\!\!\!\!&\!\!\!\!\!14.4
\!\!\!\!\!&\!\!\!\!\!2.7
\!\!\!\!\!&\!\!\!\!\!14.1
\!\!\!\!\!&\!\!\!\!\!25.3
\!\!\!\!\!&\!\!\!\!\!16.3
\!\!\!\!\!&\!\!\!\!\!37.4
\!\!\!\!\!&\!\!\!\!\!13.6
\!\!\!\!\!&\!\!\!\!\!10.8
\!\!\!\!\!&\!\!\!\!\!22.4
\!\!\!\!\!&\!\!\!\!\!26.8
\!\!\!\!\!&\!\!\!\!\!54.5
\!\!\!\!\!&\!\!\!\!\!
\!\!\!\!\!&\!\!\!\!\!\textcolor{blue}{21.3}
\\
\!\!\! {\small fc7+NN} \!\!\!\!\!&\!\!\!\!\! {Y}
\!\!\!\!\!&\!\!\!\!\!2.1
\!\!\!\!\!&\!\!\!\!\!22.6
\!\!\!\!\!&\!\!\!\!\!1.2
\!\!\!\!\!&\!\!\!\!\!1.1
\!\!\!\!\!&\!\!\!\!\!2.2
\!\!\!\!\!&\!\!\!\!\!6.1
\!\!\!\!\!&\!\!\!\!\!2.3
\!\!\!\!\!&\!\!\!\!\!8.8
\!\!\!\!\!&\!\!\!\!\!40.6
\!\!\!\!\!&\!\!\!\!\!10.6
\!\!\!\!\!&\!\!\!\!\!7.0
\!\!\!\!\!&\!\!\!\!\!5.3
\!\!\!\!\!&\!\!\!\!\!21.1
\!\!\!\!\!&\!\!\!\!\!55.9
\!\!\!\!\!&\!\!\!\!\!
\!\!\!\!\!&\!\!\!\!\!\textcolor{blue}{14.0}
\\
\!\!\! {\small fc7+sp+linearSVM} \!\!\!\!\!&\!\!\!\!\! {Y}
\!\!\!\!\!&\!\!\!\!\!16.3
\!\!\!\!\!&\!\!\!\!\!10.7
\!\!\!\!\!&\!\!\!\!\!22.0
\!\!\!\!\!&\!\!\!\!\!31.9
\!\!\!\!\!&\!\!\!\!\!4.9
\!\!\!\!\!&\!\!\!\!\!20.2
\!\!\!\!\!&\!\!\!\!\!26.3
\!\!\!\!\!&\!\!\!\!\!{\bf 23.7}
\!\!\!\!\!&\!\!\!\!\!35.3
\!\!\!\!\!&\!\!\!\!\!12.1
\!\!\!\!\!&\!\!\!\!\!12.4
\!\!\!\!\!&\!\!\!\!\!36.8
\!\!\!\!\!&\!\!\!\!\!22.4
\!\!\!\!\!&\!\!\!\!\!48.6
\!\!\!\!\!&\!\!\!\!\!
\!\!\!\!\!&\!\!\!\!\!\textcolor{blue}{25.7}
\\
\!\!\! {\small fc7+sp+RBF-SVM} \!\!\!\!\!&\!\!\!\!\! {Y}
\!\!\!\!\!&\!\!\!\!\!7.9
\!\!\!\!\!&\!\!\!\!\!27.1
\!\!\!\!\!&\!\!\!\!\!7.3
\!\!\!\!\!&\!\!\!\!\!14.4
\!\!\!\!\!&\!\!\!\!\!2.7
\!\!\!\!\!&\!\!\!\!\!14.1
\!\!\!\!\!&\!\!\!\!\!19.4
\!\!\!\!\!&\!\!\!\!\!16.3
\!\!\!\!\!&\!\!\!\!\!37.4
\!\!\!\!\!&\!\!\!\!\!13.6
\!\!\!\!\!&\!\!\!\!\!9.1
\!\!\!\!\!&\!\!\!\!\!22.4
\!\!\!\!\!&\!\!\!\!\!27.6
\!\!\!\!\!&\!\!\!\!\!55.0
\!\!\!\!\!&\!\!\!\!\!
\!\!\!\!\!&\!\!\!\!\!\textcolor{blue}{21.0}
\\
\!\!\! {\small fc7+sp+NN} \!\!\!\!\!&\!\!\!\!\! {Y}
\!\!\!\!\!&\!\!\!\!\!2.1
\!\!\!\!\!&\!\!\!\!\!22.6
\!\!\!\!\!&\!\!\!\!\!1.2
\!\!\!\!\!&\!\!\!\!\!1.1
\!\!\!\!\!&\!\!\!\!\!2.2
\!\!\!\!\!&\!\!\!\!\!6.1
\!\!\!\!\!&\!\!\!\!\!2.3
\!\!\!\!\!&\!\!\!\!\!8.8
\!\!\!\!\!&\!\!\!\!\!{\bf40.6}
\!\!\!\!\!&\!\!\!\!\!10.6
\!\!\!\!\!&\!\!\!\!\!7.0
\!\!\!\!\!&\!\!\!\!\!5.3
\!\!\!\!\!&\!\!\!\!\!21.1
\!\!\!\!\!&\!\!\!\!\!55.9
\!\!\!\!\!&\!\!\!\!\!
\!\!\!\!\!&\!\!\!\!\!\textcolor{blue}{14.0}
\\
\!\!\! {\small Fast-RCNN (1 ft)~\cite{FastRCNN}} \!\!\!\!\!&\!\!\!\!\! {N}
\!\!\!\!\!&\!\!\!\!\!3.2
\!\!\!\!\!&\!\!\!\!\!6.8
\!\!\!\!\!&\!\!\!\!\!11.0
\!\!\!\!\!&\!\!\!\!\!11.2
\!\!\!\!\!&\!\!\!\!\!1.6
\!\!\!\!\!&\!\!\!\!\!7.4
\!\!\!\!\!&\!\!\!\!\!23.0
\!\!\!\!\!&\!\!\!\!\!1.9
\!\!\!\!\!&\!\!\!\!\!2.1
\!\!\!\!\!&\!\!\!\!\!2.5
\!\!\!\!\!&\!\!\!\!\!3.8
\!\!\!\!\!&\!\!\!\!\!11.8
\!\!\!\!\!&\!\!\!\!\!14.5
\!\!\!\!\!&\!\!\!\!\!19.5
\!\!\!\!\!&\!\!\!\!\!
\!\!\!\!\!&\!\!\!\!\!\textcolor{blue}{10.0}
\\
\!\!\! {\small Fast-RCNN (2 fts)~\cite{FastRCNN}} \!\!\!\!\!&\!\!\!\!\! {Y}
\!\!\!\!\!&\!\!\!\!\!6.3
\!\!\!\!\!&\!\!\!\!\!15.3
\!\!\!\!\!&\!\!\!\!\!{\bf 39.0}
\!\!\!\!\!&\!\!\!\!\!34.6
\!\!\!\!\!&\!\!\!\!\!{\bf 36.2}
\!\!\!\!\!&\!\!\!\!\!{\bf 43.6}
\!\!\!\!\!&\!\!\!\!\!46.5
\!\!\!\!\!&\!\!\!\!\!20.5
\!\!\!\!\!&\!\!\!\!\!26.7
\!\!\!\!\!&\!\!\!\!\!13.1
\!\!\!\!\!&\!\!\!\!\!36.6
\!\!\!\!\!&\!\!\!\!\!{\bf 56.6}
\!\!\!\!\!&\!\!\!\!\!47.8
\!\!\!\!\!&\!\!\!\!\!57.3
\!\!\!\!\!&\!\!\!\!\!
\!\!\!\!\!&\!\!\!\!\!\textcolor{blue}{31.9}
\\
\!\!\! {\small Ours} \!\!\!\!\!&\!\!\!\!\! {Y}
\!\!\!\!\!&\!\!\!\!\!{\bf 37.9}
\!\!\!\!\!&\!\!\!\!\!{\bf 35.6}
\!\!\!\!\!&\!\!\!\!\!15.2
\!\!\!\!\!&\!\!\!\!\!{\bf 41.0}
\!\!\!\!\!&\!\!\!\!\!27.6
\!\!\!\!\!&\!\!\!\!\!39.9
\!\!\!\!\!&\!\!\!\!\!{\bf 53.5}
\!\!\!\!\!&\!\!\!\!\!15.8
\!\!\!\!\!&\!\!\!\!\!20.9
\!\!\!\!\!&\!\!\!\!\!{\bf 28.3}
\!\!\!\!\!&\!\!\!\!\!{\bf 55.4}
\!\!\!\!\!&\!\!\!\!\!32.9
\!\!\!\!\!&\!\!\!\!\!{\bf 51.8}
\!\!\!\!\!&\!\!\!\!\!{\bf 67.3}
\!\!\!\!\!&\!\!\!\!\!
\!\!\!\!\!&\!\!\!\!\!\textcolor{blue}{\bf 39.1}
\\
\hline
\end{tabular}}
\end{table*}

\subsubsection{Hybrid And-Or graph for semantic parts}

The explanatory graph makes it plausible to transfer middle-layer patterns from CNNs to semantic object parts. In order to test the transferability of patterns in the explanatory graph, we introduce a further extension of the disentangling graph, \emph{i.e.} using a hybrid And-Or graph (AOG) to associate part patterns in the explanatory graph with explicit part names. The structure of the AOG is inspired by \cite{NineAOG}, and the learning of the AOG was originally proposed in \cite{CNNAoG}. We briefly introduce basic inference logic and settings of the AOG as follows.

As shown in Fig.~\ref{fig:hybrid}, the AOG encodes a four-layer hierarchy for each semantic part, \emph{i.e.} the semantic part (OR node), part templates (AND node), latent patterns (OR nodes, those from the explanatory graph), and neural units (terminal nodes).
\begin{center}
\resizebox{\linewidth}{!}{\begin{tabular}{c|lcc}
\hline
Layer \!\!&\!\! Name \!\!&\!\! Node type \!\!&\!\! Notation\\
\hline
1 \!\!&\!\! semantic part \!\!&\!\! OR node \!\!&\!\! $V^{\textrm{sem}}$\\
2 \!\!&\!\! part template \!\!&\!\! AND node \!\!&\!\! $V^{\textrm{tmp}}\!\in\!\Omega^{\textrm{tmp}}$\\
3 \!\!&\!\! latent pattern \!\!&\!\! OR node \!\!&\!\! $V^{\textrm{lat}}\!\in\!\Omega^{\textrm{lat}}$\\
4 \!\!&\!\! neural unit \!\!&\!\! Terminal node \!\!&\!\! $x\!\in\!\Omega^{\textrm{unt}}$\\
\hline
\end{tabular}}
\end{center}
where latent patterns correspond to nodes from the explanatory graph.

In the AOG, each OR node (\emph{e.g.} a semantic part or a latent pattern) contains a list of alternative appearance (or deformation) candidates. Each AND node (\emph{e.g.} a part template) uses a number of latent patterns to describe its compositional regions.
\begin{itemize}
\item The OR node of a semantic part contains a total of $m$ part templates to represent alternative appearance or pose candidates of the part.
\item Each part template (AND node) retrieve $K$ patterns from the explanatory graph as children. These patterns describe compositional regions of the part.
\item Each latent pattern (OR node) has all units in its corresponding filter's feature map as children, which represent its deformation candidates on image $I$.
\end{itemize}

\textbf{Technical details:} Based on the AOG, we use the extracted patterns to infer semantic parts in a bottom-up manner. We first compute inference scores of different units at the bottom layer \emph{w.r.t.} different patterns, and then we propagate inference scores up to the layers of part templates and the semantic part for part localization.

The top OR node of the semantic part $V^{\textrm{sem}}$ contains a total of $m$ part templates to represent alternative appearance or pose candidates of the part. We manually define the composition of the $M$ part templates. During part-inference process, given an image $I$, $V^{\textrm{sem}}$ selects its best child as the true part template:
\begin{equation}
\begin{split}
S_{V^{\textrm{sem}}}&=\max_{V^{\textrm{tmp}}\in Child(V^{\textrm{sem}})}S_{V^{\textrm{tmp}}}\\
{\bf p}_{V^{\textrm{sem}}}&={\bf p}_{\hat{V}^{\textrm{tmp}}}
\end{split}
\end{equation}
where $S_{V^{X}}, X\in\{\textrm{sem,tmp,lat,unit}\}$ denotes the inference score of $V^{X}$.

Then, each part template $V^{\textrm{tmp}}$ uses a number of latent patterns to describe sub-regions of the part. In the scenario of one-shot learning, we only annotate one part sample belonging to the part template. Then, we retrieve patterns that are related to the annotated part from all nodes in the disentangling graph. Given the inference score $S_{V^{\textrm{lat}}}$ and inferred position ${\bf p}_{V^{\textrm{lat}}}$ of each latent pattern $V^{\textrm{lat}}$ on $I$, we retrieve the top $K$ latent patterns with the highest scores of $S_{V^{\textrm{lat}}}{\mathcal N}({\bf p}_{V^{\textrm{lat}}}|\mu={\bf p}_{V^{\textrm{tmp}}}^{*},\sigma^2)$ as children of $V^{\textrm{tmp}}$. ${\bf p}_{V^{\textrm{tmp}}}^{*}$ denotes the annotated position of the part $V^{\textrm{tmp}}$; $\sigma^2=(0.3\times\!image\,width)^2$ is a constant variation.

When we have extracted a set of latent patterns for a part template, given a new image, we can use inference results of the latent patterns to localize the part template:
\begin{equation}
\begin{split}
S_{V^{\textrm{tmp}}}&=\sum_{V^{\textrm{lat}}\in Child(V^{\textrm{tmp}})}S_{V^{\textrm{lat}}}\\
{\bf p}_{V^{\textrm{tmp}}}&=\underset{V^{\textrm{lat}}\in Child(V^{\textrm{tmp}})}{\textrm{mean}}\big\{{\bf p}_{V^{\textrm{lat}}}+\Delta{\bf p}_{V^{\textrm{lat}},V^{\textrm{tmp}}}\big\}
\end{split}
\end{equation}
where $\Delta{\bf p}_{V^{\textrm{lat}},V^{\textrm{tmp}}}$ denotes a constant displacement from $V^{\textrm{lat}}$ to $V^{\textrm{tmp}}$.

Each latent pattern $V^{\textrm{lat}}$ has a channel of units as children, which represent its deformation candidates on image $I$. The score of each unit $x$ is given as {$S_{V^{\textrm{lat}}\rightarrow x}=F(x)P({\bf p}_{x}|V^{\textrm{lat}},{\bf R}_{L+1},{\boldsymbol\theta}_{L})$}. The OR node of $V^{\textrm{lat}}$ selects the unit with the maximum score as its deformation configuration:
\begin{equation}
\begin{split}
S_{V^{\textrm{lat}}}&={\max}_{x:V^{\textrm{lat}}\in\Omega_{L,d_{x}}}\,S_{V^{\textrm{lat}}\rightarrow x}\\
{\bf p}_{V^{\textrm{lat}}}&={\bf p}_{\hat{x}}
\end{split}
\end{equation}
Please see \cite{CNNAoG} for details of the AOG.

\subsubsection{Experimental settings of three-shot learning}

Given a fine-tuned VGG-16 network, we learned an explanatory graph and built the AOG upon the explanatory graph following the scenario of few-shot learning in \cite{CNNAoG}. For each category, we set three templates for the head part ($m=3$) and used three part-box annotations for the three templates. Note that we used object images without part annotations to learn the explanatory graph, and we used three part annotations provided by \cite{CNNAoG} for each part to build the AOG. All these object-box annotations and part annotations were equally provided to all baselines to enable fair comparisons (besides part annotations, all baselines also used object annotations contained in the datasets for learning). We set $K=0.1\sum_{L,d}N_{L,d}$ to learn AOGs for categories in the ILSVRC Animal-Part and CUB200 datasets and set $K=0.4\sum_{L,d}N_{L,d}$ for VOC Part categories. Then, we used the AOGs to localize semantic parts on objects.

\textbf{Baselines:}{\verb| |} We compared AOGs with a total of fourteen baselines for part localization. The baselines included (i) approaches for object detection (\emph{i.e.} directly detecting target parts from objects), (ii) graphical/part models for part localization, and (iii) the methods selecting CNN patterns to describe object parts.

The first baseline was the standard fast-RCNN~\cite{FastRCNN}, namely \textit{Fast-RCNN (1 ft)}, which directly fine-tuned a VGG-16 network based on part annotations. Then, the second baseline, namely \textit{Fast-RCNN (2 fts)}, first used massive object-box annotations in the target category to fine-tune the VGG-16 network with the loss of object detection. Then, given part annotations, Fast-RCNN (2 fts) further fine-tuned the VGG-16 to detect object parts. We used \cite{CNNSemanticPart} as the third baseline, namely \textit{CNN-PDD}. CNN-PDD selected certain filters of a CNN to localize the target part. In CNN-PDD, the CNN was pre-trained using the ImageNet dataset~\cite{ImageNet}. Just like Fast-RCNN (2 ft), we extended \cite{CNNSemanticPart} as the fourth baseline \textit{CNN-PDD-ft}, which fine-tuned a VGG-16 network using object-box annotations before applying the technique of \cite{CNNSemanticPart}. The fifth and sixth baselines were DPM-related methods, \emph{i.e.} the strongly supervised DPM (\textit{SS-DPM-Part})~\cite{SSDPM} and the technique in \cite{PLDPM} (\textit{PL-DPM-Part}), respectively. Then, the seventh baseline, namely \textit{Part-Graph}, used a graphical model for part localization~\cite{SemanticPart}. For weakly supervised learning, ``simple'' methods are usually insensitive to model over-fitting. Thus, we designed six baselines as follows. First, we used object-box annotations in a category to fine-tune the VGG-16 network. Then, given a few well-cropped object images, we used the selective search~\cite{SelectiveSearch} to collect image patches, and used the VGG-16 network to extract \textit{fc7} features from these patches. The baselines \textit{fc7+linearSVM}, \textit{fc7+RBF-SVM}, \textit{fc7+NN} used a linear SVM, an RBF-SVM, and the nearest-neighbor method (selecting the patch closest to the annotated part), respectively, to detect the target part. The other three baseline \textit{fc7+sp+linearSVM}, \textit{fc7+sp+RBF-SVM}, \textit{fc7+sp+NN} combined both the \textit{fc7} feature and the spatial position {\small$(x,y)$ ($-1\leq x,y\leq1$)} of each image patch as features for part detection. The last competing method is weakly supervised mining of part patterns from CNNs~\cite{CNNAoG}, namely \textit{supervised-AOG}. Unlike our method (unsupervised), \textit{supervised-AOG} used part annotations to extract part patterns.

\begin{table}[t]
\centering
\caption{Normalized distance of part localization. We compared supervised and unsupervised mining of part patterns.}
\label{tab:cnnaog}
\resizebox{1.0\linewidth}{!}{\begin{tabular}{l|ccc}
\hline
{Dataset} & {ILSVRC DET} & {VOC} & {CUB200}\\
 & {Animal} & {Part} & {-2011}\\
\hline
{Supervised-AOG} & 0.1344 & 0.1767 & 0.0915\\
{Ours (unsupervised)} & {\bf0.1250} & {\bf0.1765} & {\bf0.0862}\\
\hline
\end{tabular}}
\end{table}

\textbf{Comparisons:}{\verb| |} We divided all baselines into three groups. The first group, namely \textit{not-learn parts}, included traditional methods without using deep features, such as SS-DPM-Part, PL-DPM-Part, and Part-Graph. These methods did not learn deep features\footnote[5]{Representation learning in these methods only used object-box annotations, which is independent to part annotations. A few part annotations were used to select off-the-shelf pre-trained features.}. The second group, termed \textit{super-learn parts}, contained Fast-RCNN (1 ft), Fast-RCNN (2 ft), CNN-PDD, CNN-PDD-ft, supervised-AOG, fc7+linearSVM, and fc7+sp+linearSVM. These methods learned deep features using part annotations, \emph{e.g.} fast-RCNN methods used part annotations to learn features; supervised-AOG used part annotations to select filters from CNNs to localize parts. The third group (\textit{unsuper-learn parts}) included CNN-PDD, CNN-PDD-ft, and our method. These methods learned deep features using object-level annotations, rather than part annotations.

\begin{figure*}[t]
\centering
\includegraphics[width=0.8\linewidth]{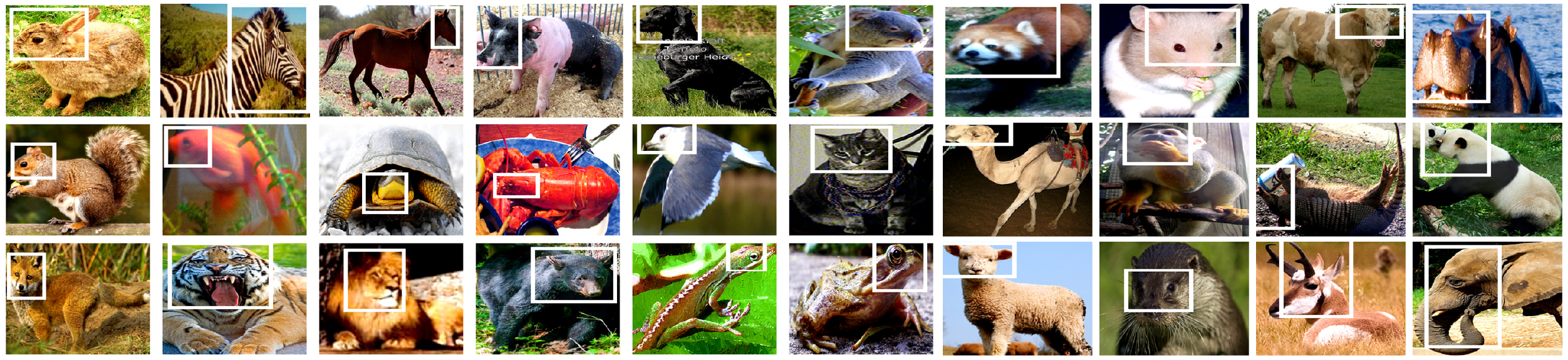}
\caption{Localization results based on AOGs that are learned using three annotations of the head part.}
\label{fig:localization}
\end{figure*}

Fig.~\ref{fig:localization} visualizes localization results based on AOGs, which were learned using three annotations of the head part of each category. We used the normalized distance (used in \cite{CNNAoG,CNNSemanticPart}) and the traditional intersection-over-union (IoU) criterion to evaluate the localization performance. Tables~\ref{tab:CUB}, \ref{tab:VOC}, \ref{tab:VOC_IOU}, \ref{tab:imgnet}, and \ref{tab:imgnet_IOU} show part-localization results on the CUB200-2011 dataset~\cite{CUB200}, the VOC Part dataset~\cite{SemanticPart}, and the ILSVRC 2013 DET Animal-Part dataset~\cite{CNNAoG}. AOGs based on our graph nodes exhibited outperformed all baselines in few-shot learning. Note that our AOGs simply localized the center of an object part without sophisticatedly modeling the scale of the part. Thus, detection-based methods, which also estimated the part scale, performed better in very few cases. Table~\ref{tab:cnnaog} compares the unsupervised and supervised learning of neural patterns. In the experiment, our method outperformed all baselines, even including approaches that learned part features using part annotations.

\section{Conclusion and discussions}

In this paper, we have developed a simple yet effective method to learn an explanatory graph that reveals knowledge hierarchy inside conv-layers of a pre-trained CNN. The explanatory graph can be regarded as a concise and meaningful summarization of CNN knowledge in intermediate layers, which filters out noisy activations, disentangles part patterns from each filter, and models co-activation relationships and spatial relationships between part patterns. Experiments showed that our patterns had significantly higher stability than baselines. More crucially, our method can be applied to different types of networks, including the VGG-16, residual networks, and the VAE-GAN, to explain their conv-layers.

The transparent representation of the explanatory graph boosts the transferability of CNN features. Part-localization experiments well demonstrated the good transferability of CNN patterns in graph nodes. Our method even outperformed the supervised learning of part representations. Nevertheless, the explanatory graph is just a rough representation of CNN knowledge. It is still difficult to well disentangle textural patterns from filters of the CNN.

\ifCLASSOPTIONcompsoc
  \section*{Acknowledgments}
\else
  \section*{Acknowledgment}
\fi
This work is supported by ONR MURI project N00014-16-1-2007, DARPA XAI Award N66001-17-2-4029, and NSF IIS 1423305.

\ifCLASSOPTIONcaptionsoff
  \newpage
\fi

\bibliographystyle{IEEE}
\bibliography{TheBib}

\vspace{-30pt}
\begin{IEEEbiography}[{\includegraphics[width=1in,height=1.25in,clip,keepaspectratio]{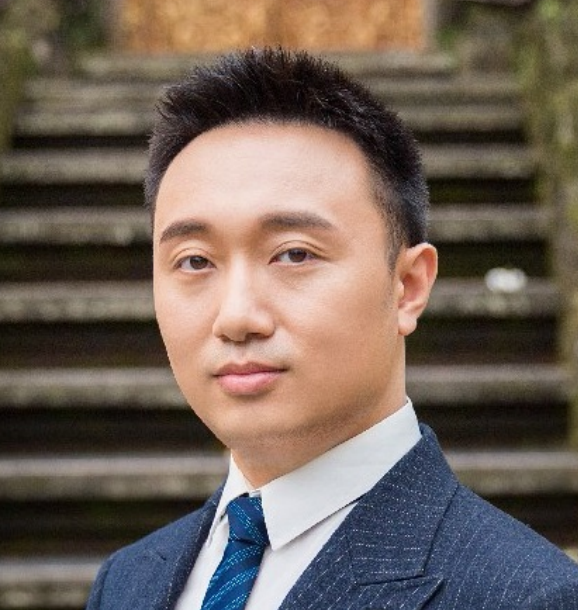}}]{Quanshi Zhang}
received the B.S. degree in machine intelligence from Peking University, China, in 2009 and M.S. and Ph.D. degrees in center for spatial information science from the University of Tokyo, Japan, in 2011 and 2014, respectively. In 2014, he went to the University of California, Los Angeles, as a post-doctoral associate. Now, he is an associate professor at the Shanghai Jiao Tong University. His research interests include computer vision, machine learning, and robotics.
\end{IEEEbiography}

\vspace{-30pt}
\begin{IEEEbiography}[{\includegraphics[width=1in,height=1.25in,clip,keepaspectratio]{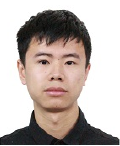}}]{Xin Wang} is an internship student at the Shanghai Jiao Tong University.
\end{IEEEbiography}

\vspace{-30pt}
\begin{IEEEbiography}[{\includegraphics[width=1in,height=1.25in,clip,keepaspectratio]{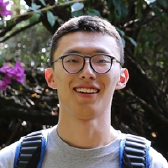}}]{Ruiming Cao}
received the B.S. degree in computer science from the University of California, Los Angeles, in 2017. Now, he is a master student at the University of California, Los Angeles. His research mainly focuses on computer vision.
\end{IEEEbiography}

\vspace{-30pt}
\begin{IEEEbiography}[{\includegraphics[width=1in,height=1.25in,clip,keepaspectratio]{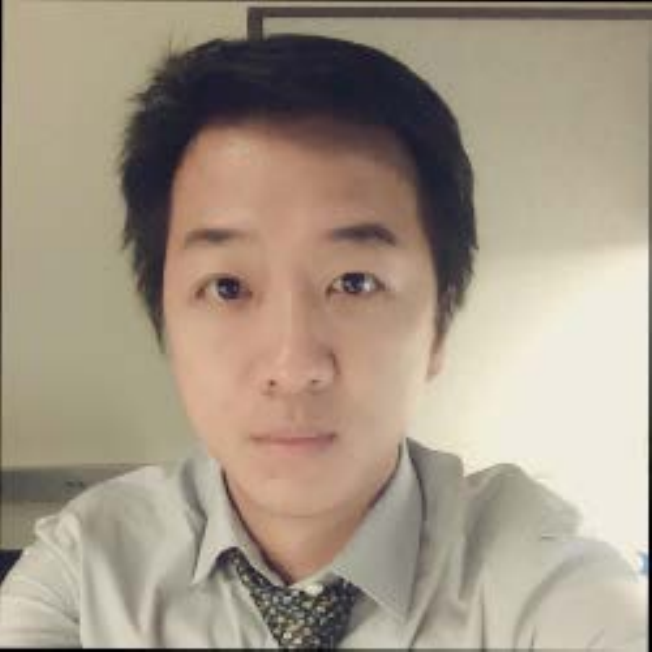}}]{Feng Shi} is a Ph.D. student at the University of California, Los Angeles. His research mainly focuses on computer vision and electric engineering.
\end{IEEEbiography}

\vspace{-30pt}
\begin{IEEEbiography}[{\includegraphics[width=1in,height=1.25in,clip,keepaspectratio]{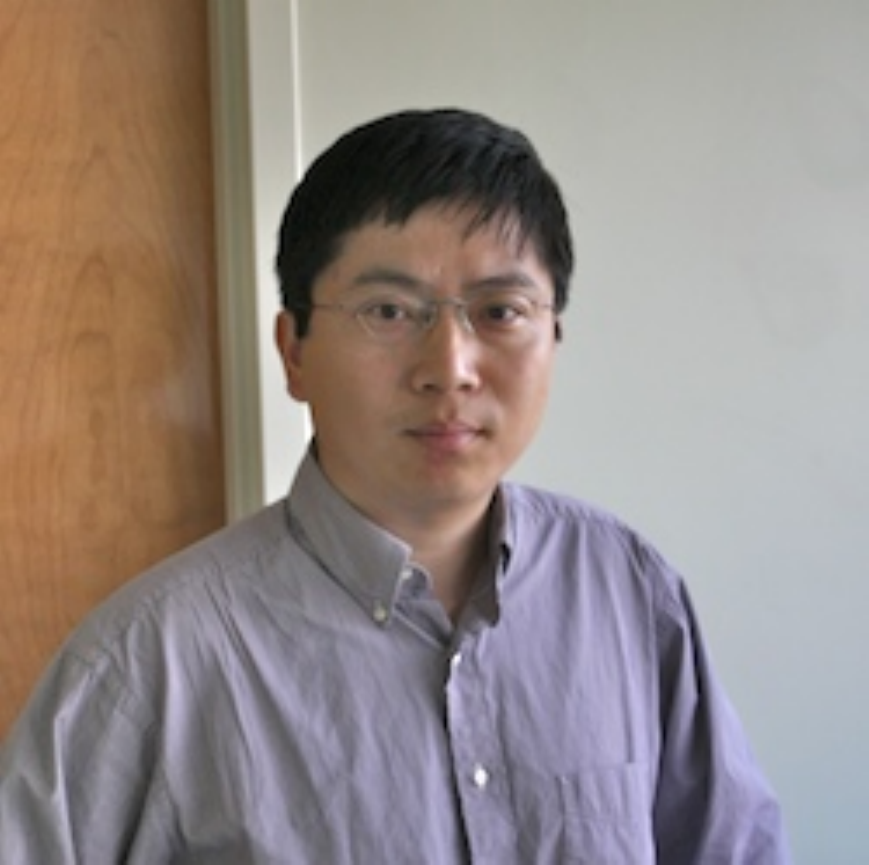}}]{Ying Nian Wu}
received a Ph.D. degree from the Harvard University in 1996. He was an Assistant Professor at the University of Michigan between 1997 and 1999 and an Assistant Professor at the University of California, Los Angeles between 1999 and 2001. He became an Associate Professor at the University of California, Los Angeles in 2001. From 2006 to now, he is a professor at the University of California, Los Angeles. His research interests include statistics, machine learning, and computer vision.
\end{IEEEbiography}

\vspace{-30pt}
\begin{IEEEbiography}[{\includegraphics[width=1in,height=1.25in,clip,keepaspectratio]{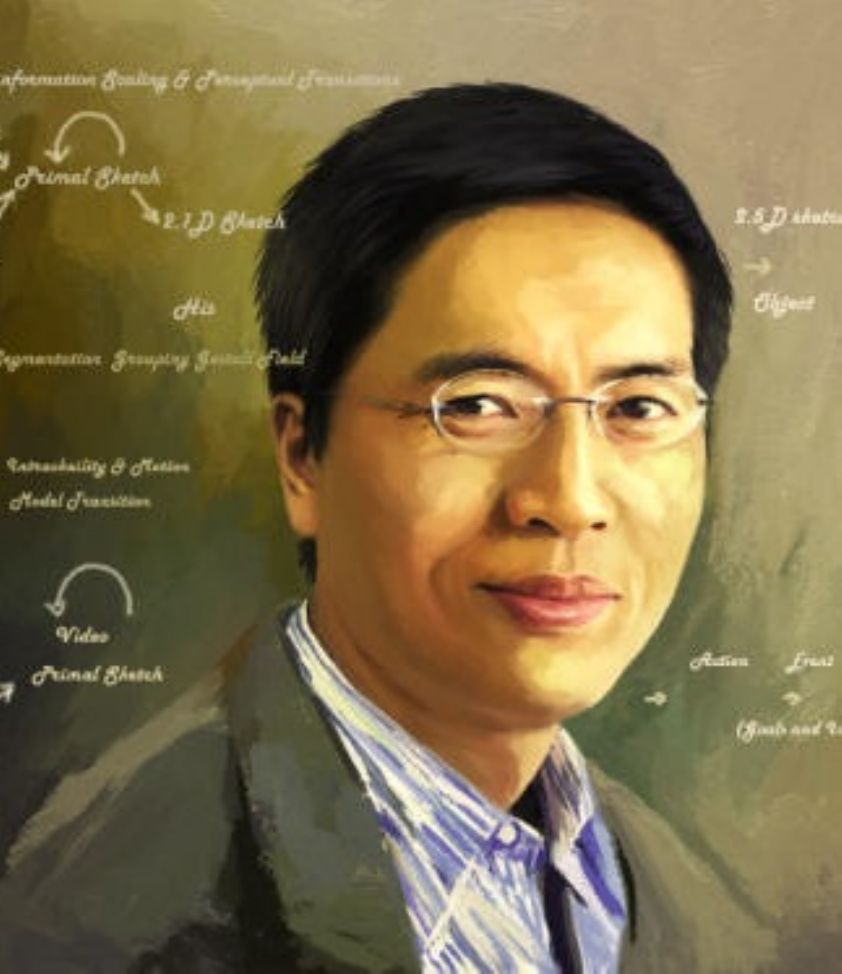}}]{Song-Chun Zhu} received a Ph.D. degree from Harvard University, and is a professor with the Department of Statistics and the Department of Computer Science at UCLA. His research interests include computer vision, statistical modeling and learning, cognition and AI, and visual arts. He received a number of honors, including the Marr Prize in 2003 with Z. Tu et. al. on image parsing,the Aggarwal prize from the Int'l Association of Pattern Recognition in 2008, twice Marr Prize honorary nominations in 1999 for texture modeling and 2007 for object modeling with Y.N. Wu et al., a Sloan Fellowship in 2001, the US NSF Career Award in 2001, and the US ONR Young Investigator Award in 2001. He is a Fellow of IEEE.
\end{IEEEbiography}
\vfill

\end{document}